\documentclass{article}
\usepackage{iclr2025_conference,times}

\usepackage{amsmath,amsfonts,bm}

\def\eqref#1{equation~\ref{#1}}
\def\Eqref#1{Equation~\ref{#1}}

\def\1{\bm{1}}

\DeclareMathAlphabet{\mathsfit}{\encodingdefault}{\sfdefault}{m}{sl}
\SetMathAlphabet{\mathsfit}{bold}{\encodingdefault}{\sfdefault}{bx}{n}

\def\gD{{\mathcal{D}}}
\def\gE{{\mathcal{E}}}

\def\gG{{\mathcal{G}}}

\def\gN{{\mathcal{N}}}

\newcommand{\E}{\mathbb{E}}

\usepackage{hyperref}
\usepackage{url}
\usepackage[pdftex]{graphicx}
\usepackage{subcaption}
\usepackage{multirow}
\usepackage{mathtools}
\usepackage[normalem]{ulem}

\title{Lost-in-Distance: Impact of Contextual Proximity on LLM Performance in Graph Tasks}

\author{Hamed Firooz, Maziar Sanjabi, Wenlong Jiang\thanks{work is done while at LinkedIn FAIT} , \& Xiaoling Zhai \\
Foundation AI Technologies (FAIT)\\
LinkedIn\\
\texttt{\{hfirooz, maz, wenjiang, xzhai\}@linkedin.com}
}

\arxivcopy 

\begin{document}

\maketitle

\begin{abstract}
Despite significant advancements, Large Language Models (LLMs) exhibit blind spots that impair their ability to retrieve and process relevant contextual data effectively. We demonstrate that LLM performance in graph tasks with complexities beyond the ``needle-in-a-haystack" scenario—where solving the problem requires cross-referencing and reasoning across multiple subproblems \emph{jointly}—is influenced by the proximity of relevant information within the context, a phenomenon we term ``lost-in-distance". We examine two fundamental graph tasks: identifying common connections between two nodes and assessing similarity among three nodes, and show that the model's performance in these tasks significantly depends on the relative positioning of common edges. We evaluate three publicly available LLMs using various graph encoding techniques that represent graph structures for LLM input. We propose a formulation for the lost-in-distance phenomenon and demonstrate that lost-in-distance and lost-in-the-middle phenomena occur independently. Results indicate that model accuracy can decline by up to 6x as the distance between node connections increases, independent of graph encoding and model size. 
\end{abstract}

\section{Introduction}\label{s:intro}
Large Language Models (LLMs) have attained an unprecedented level of generality by leveraging scale and attention-based architectures \citep{kaplan2020scaling, vaswani2017attention}. These models exhibit remarkable, often superhuman, capabilities across a diverse range of tasks, including language translation, reading comprehension, and question answering \citep{costa2022no, sanh2021multitask}. Additionally, LLMs are increasingly serving as essential and flexible building blocks for various user-facing machine learning and artificial intelligence applications beyond traditional language processing domains, such as recommendation systems \citep{geng2022recommendation}, graph-related tasks \citep{wang2024can}, knowledge bases \citep{alkhamissi2022review, petroni2019language}, and more. These applications highlight the versatility of LLMs but also expose new challenges in handling domain-specific data encoded as textual input.

Particularly, by leveraging the extensive common knowledge and powerful semantic comprehension abilities of LLMs, recent research has aimed to apply them to tasks related to graph structures \citep{wang2024can}. LLMs are increasingly being adopted for a variety of tasks that involve graph structures, such as planning in robotics \citep{andreas2022language, huang2022language}, knowledge extraction using knowledge graphs \citep{shen2020exploiting, saxena2020improving, sun2023think}, and multi-hop question answering \citep{creswell2022selection, fang2019hierarchical}. 

Building upon these applications, \citet{chen2024graphwiz} and \citet{luo2024graphinstruct} introduce GraphWiz and GraphLM, respectively—fine-tuned LLMs on graph instruction and reasoning tasks—which exhibit superior performance compared to general large language models in graph understanding and reasoning capabilities. Recent works by \citet{sanford2024understanding}, \citet{perozzi2024let}, and \citet{agarwal2020knowledge} have demonstrated that graph tasks can be encoded into textual formats, enabling general pre-trained LLMs to solve them as out-of-domain tasks. This innovative approach effectively transforms graph problems into a language that LLMs can understand and process.

While general LLMs are being expanded in many applications including graph understanding, they suffer from certain blind spots that significantly affect their performance. In particular, how these models process information in their prompt and retrieve relevant data to solve the task at hand remains an active area of research \citep{kaddour2023challenges}. Understanding these limitations is crucial for extending context length \citep{team2023gemini, xu2023retrieval, chen2023extending} and improving in-context learning \citep{zhou2022large, wei2023larger, an2024make}.  Recent works have shown that the performance of LLMs depends on the location of information in their prompt. Primarily, \citet{liu2023lost} introduces the ``lost-in-the-middle" phenomenon, where information placed in the middle of a prompt is less effectively utilized by the model compared to information at the beginning or end, resulting in significant performance degradation when the position of relevant information in the context changes.

Unlike previous works that primarily focus on the limitations of LLMs in natural language processing tasks, this study investigates the deficiencies of these models in tasks beyond NLP, focusing on solving fundamental graph problems. While leveraging LLMs to address graph-related problems has broad practical applications~\citep{perozzi2024let,colon2021combining,xie2023graph}, recent studies have shown that the sequential order in which a graph is described significantly affects LLMs' comprehension of graph structures and their performance in solving graph tasks~\citep{ge2024can}. In this study, we aim to explore the underlying phenomena and blind spots that contribute to performance variability based on the ordered description of graphs. Our analysis reveals that LLMs not only have blind spots regarding the locations of information within the prompt but that their ability to solve complex tasks also depends on the \emph{relative} positioning of this information. Since graph tasks require understanding graph structure and the relationships between nodes and edges, the sequential arrangement of data plays a crucial role in how well LLMs can integrate and reason over that information. These insights suggest that improvements in graph comprehension by LLMs must account for both the placement and relative distance of information to enhance performance in graph-related problem-solving.

Particularly, we look into Common Connection and Similarity tasks, which are the main algorithms used as the backbone of many applications such as molecular design \citep{tan2023target, xia2023mole}, social network analysis \citep{gao2024graph}, and recommendation systems \citep{li2023survey}. For example, these tasks are the main algorithms in ``user-user" and ``user-item" recommendations in large industry recommendation products \citep{xie2023graph, huang2015social, wu2022graph, dai2024enhancing}. These tasks not only require understanding of subgraph structures but also demand integration of information and reasoning across subgraphs. We demonstrate that strong, publicly available LLMs universally degrade in performance as  when relevant pieces of information are distant from each other. Our analysis shows that this effect is present even when one controls for the effects absolute position of the relevant information in the prompt. To summarize
\begin{itemize}
    \item For tasks that require cross-subgraph information lookup, such as identifying common connections or measuring similarity, model performance not only degrades due to the ``lost-in-the-middle" effect based on the absolute positions but is also affected by the relative distance between pieces of information in the context—a phenomenon we term ``lost-in-distance". The further apart the information is, the more the model's performance deteriorates.

    \item We demonstrate these shortcomings across different graph encoding algorithms and various publicly available LLMs such as Llama-3-8B, Llama-3-70B \citep{dubey2024llama} and GPT-4 \citep{achiam2023gpt} indicating a universal limitation in current architectures.

    \item We formulate the goodness of fit for the lost-in-distance phenomenon and demonstrate its compounded impact when combined with the lost-in-the-middle phenomenon in solving graph-related tasks.
\end{itemize}

Our findings suggest that current LLMs have inherent limitations in processing contextual information that is not sequentially localized or is widely dispersed within the input. This has significant implications for their application in domains that require complex reasoning over structured data, such as graph analysis.

\subsection{Notations and Definitions}
We define a graph $\mathcal{G} = (\mathcal{V}, \mathcal{E})$, where $\mathcal{V} = \{v_{1}, v_{2}, \dots, v_{n}\}$ and $\mathcal{E}$ represent the sets of nodes and edges, respectively. If nodes $v_{i}$ and $v_{j}$ are directly connected, we denote the edge between them as $e_{ij} \in \mathcal{E}$. The neighbors of node $v_{i}$ are defined as $\mathcal{N}(v_{i}) = \{v_{k} \in \mathcal{V} \mid e_{ik} \in \mathcal{E}\}$. A subgraph associated with node $v_{i}$ is defined as $\mathcal{G}_{v_{i}} = (\{v_{i}\} \cup \mathcal{N}(v_{i}), \mathcal{E}_{v_{i}})$, where $\mathcal{E}_{v_{i}} = \{e_{ij} \mid e_{ij} \in \mathcal{E},\ v_{j} \in \mathcal{N}(v_{i})\}$.

We define the distance between a common node $v$ within two subgraphs $\mathcal{G}_u$ and $\mathcal{G}_z$ as the number of tokens separating the two occurrences of node $v$ in the context (i.e., the textual representation of the subgraphs). The overall distance between relevant information for common connections between the two subgraphs is defined as the median of all such distances computed for each common node. Throughout the paper, we use $p$ to indicate position and $d$ to indicate distance.

We use accuracy, as defined below, to measure the performance of an LLM model in solving a given task:
\begin{equation}\label{math:accuracy}
\mathrm{Accuracy} = \frac{1}{N} \sum_{i=1}^{N} \mathbf{1}_{\{y_{i} = \hat{y}_{i}\}} \times 100\%,
\end{equation}
where $N$ is the total number of samples in the task, and $y_{i}$ and $\hat{y}_{i}$ denote the true answer and the model's answer for the $i$th sample, respectively. If the output of the LLM for sample $i$ is degenerate—such as not following instructions or hallucinating—we consider it an incorrect answer, i.e., $y_{i} \neq \hat{y}_{i}$.

\section{Graph Encoding and Graph Tasks}\label{graph_encoding}

\subsection{Graph Encoding for LLM}
Representing graph-structured data as text is an important step in enabling LLMs to understand graph structures and provide accurate answers to questions. Encoding graphs as text involves representing both nodes and edges. Different graph encodings can lead to varying performance of LLMs in graph reasoning tasks \citep{agarwal2020knowledge, fatemi2024talk, zhang2024graphtranslator}. In this work, we encode nodes as integers, where each node is represented by a unique integer, such that $v_{i} \in \{0, 1, \dots, n-1\}$.

\begin{figure}[ht]
\begin{center}
\includegraphics[width=0.6\linewidth]{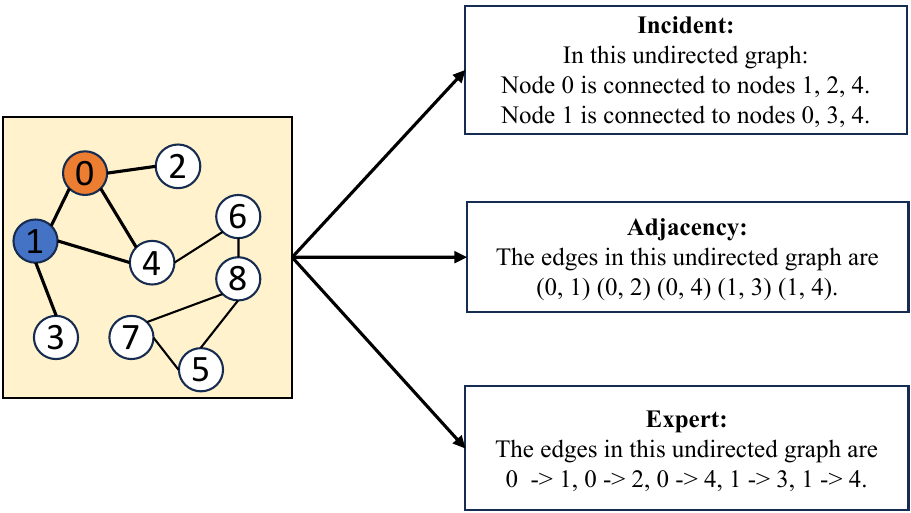}
\end{center}
\caption{Three graph encoding functions, with node $0$ and node $1$ serving as the nodes of interest. The figure is inspired by \cite{fatemi2024talk}.}
\label{graph_encoding_fig}
\end{figure}

We experiment with three encoding functions from \citet{fatemi2024talk} to encode edges in the graph, investigating whether patterns are consistently observed across different encoding functions. 
More specifically, we consider the following edge encoding functions:
\begin{itemize}
    \item \textbf{Incident:} Given a source node $v_{i}$, the edge information for node $v_{i}$ is encoded as an adjacency list in natural language. 
    For example,  ``node $v_{i}$ is connected to nodes $v_{j}, v_{k}$".
    \item \textbf{Adjacency:} Given a source node $v_{i}$ and a target node $v_{j}$, the edge is encoded as $(v_{i},v_{j})$.
    \item \textbf{Expert:} Given a source node $v_{i}$ and a target node $v_{j}$, the edge is encoded as $v_{i} \rightarrow v_{j}$. 
\end{itemize}
Since the graph tasks considered in this paper only require access to the subgraph and the subgraph structure, we encode only the edge information for the nodes of interest. This is a common practice where a subgraph is extracted from a database before being processed by a compute engine \citep{shao2013trinity}.
Figure~\ref{graph_encoding_fig} shows an example about only including a subgraph with three encoding functions in the prompt.
In this example, node $0$ and node $1$ are nodes of interest so we only encode their subgraph in the prompt.

\subsection{Graph Generation}\label{section:graph_generation}
In this paper, we build upon previous studies \citep{huang2022language, fatemi2024talk, zhang2024llm4dyg} by conducting experiments on randomly generated graphs. We utilize the Erdős–Rényi (ER) graph generator \citep{erd1959random} to create undirected graphs. We experiment with relatively large graphs comprising $n=1000$ nodes. The undirected edge $e_{ij}$ between nodes $v_{i}$ and $v_{j}$ is generated with probability $P(e_{ij} \in \gE)$. We set $P(e_{ij} \in \gE) = 0.1$ throughout the main manuscript, and results for other values of $P(e_{ij} \in \gE)$ are presented in the Appendix for brevity.

\subsection{Graph Tasks}
We aim to analyze the performance of LLMs in three fundamental graph problems which require models to have thorough understanding of the input graph structure.
\begin{enumerate}
    \item \textbf{Edge Existence:} Given two nodes $v_{i}$ and $v_{j}$ sampled from a graph $\gG$, node $v_{i}$ and node $v_{j}$ are directly connected if $e_{ij} \in \gE$. 
    The edge existence task is to ask LLMs whether node $v_{i}$ and node $v_{j}$ are directly connected. 
    \item \textbf{Common Connection:} Given two nodes $v_{i}$ and $v_{j}$ sampled from a graph $\gG$, the common connection between two nodes are $\gN (v_{i}) \cap \gN (v_{j})$.
    For this task, we ask LLMs to find the number of common connections between node $v_{i}$ and node $v_{j}$, denoted as $|\gN (v_{i}) \cap \gN (v_{j})|$.
    \item \textbf{Similarity:} Given three nodes $v_{i}$, $v_{j}$ and $v_{k}$ sampled from a graph $\gG$, we let $v_{j}$ be the source node and $v_{i}$ and $v_{k}$ be the target nodes.
    The task for LLMs is to compare the number of common connections $|\gN (v_{i}) \cap \gN (v_{j})|$ and $|\gN (v_{j}) \cap \gN (v_{k})|$.
\end{enumerate}
Note that these tasks are roughly ordered in terms of general complexity.
For example, solving the edge existence only depends on the model being able to retrieve the edge information from the representation. One step further, in finding the number of common connections, models needs to first identify the set of shared connections between two nodes and then calculate the size of that set. Finally, the similarity task is more complex than the common connection task, as it requires LLMs to consider \emph{three} nodes and identify two sets of common connections and then compare their sizes.
As a result, these tasks are a good representative set to evaluate LLMs since they require LLMs to both retrieve and reason about the graph information. Furthermore, these tasks are also essential and the building blocks for solving practical problems in applications such as recommendation systems \citep{ying2018graph}, protein folding \citep{strokach2020fast}, bad actor detection \citep{papegnies2017graph} or any other task that requires graph understanding.

\section{Lost-in-the-middle for Edge Existence}\label{s:lost_in_middle_edge_existence}
The edge existence task is analogous to the needle-in-a-haystack problem \citep{ivgi2023efficient} and the document question-answering task \citep{liu2023lost}, as it requires the LLM to retrieve the answer from the prompt without performing any computation. Building upon prior work in the literature by \citet{liu2023lost}, this study demonstrates the impact of the position of relevant information on the performance of LLMs. Specifically, it is shown that the accuracy in the edge existence task decreases when the information about the edge in question is placed in the middle of the prompt.

The prompt structure is constructed using the following procedure, which enables controlling the location of information within the prompt:

\begin{enumerate}
    \item Randomly sample two nodes from a graph along with their corresponding connections.
    \item Randomly select nine additional subgraphs and incorporate their textual encodings into the prompt. This step is necessary to examine the impact of the position of relevant information.
    \item Group the subgraph structures of the two nodes of interest and position them at the beginning, middle, or end of the input context.
    \item Query the model to determine whether an edge exists between the two nodes of interest.
\end{enumerate}

\begin{figure}[ht]
\begin{center}
\includegraphics[width=\linewidth]{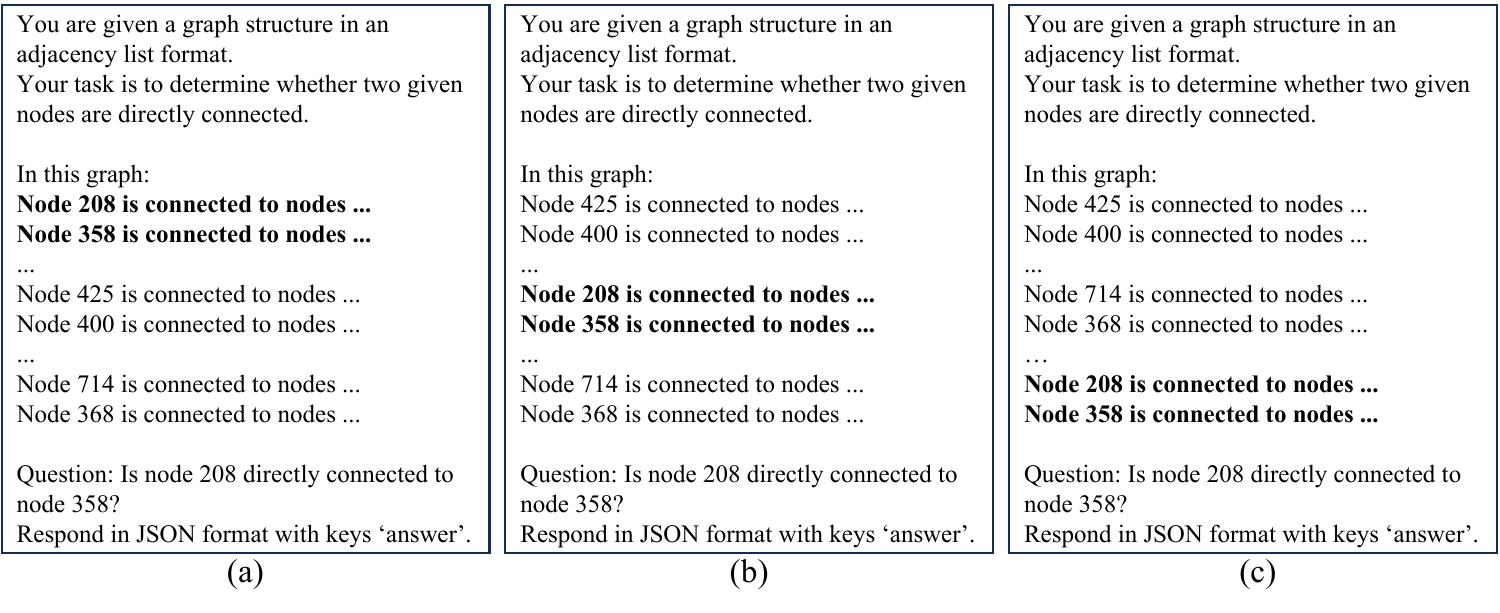}
\end{center}
\vspace{-0.2cm}
\caption{Example of the edge existence task, illustrating the placement of the nodes of interest subgraph (nodes $208$ and $358$) at (a) the beginning, (b) the middle, and (c) the end of the graph structure.}
\label{fig:edge_existence_prompt}
\end{figure}

An example of a prompt with different positions for the two nodes of interest is illustrated in Figure~\ref{fig:edge_existence_prompt}.

\subsection{Experimental Results}
\textbf{Lost-in-the-middle can happen in the edge existence task.}
To demonstrate the lost-in-the-middle phenomena in edge existence task, we experiment with one of the state of the art model as of writing this paper GPT-4 that does not use function calling. Since the lost-in-the-middle phenomenon has been extensively studied in the literature, particularly in the context of language tasks, we limit our experimentation only to GPT-4 and use its result later to study the "lost-in-distance" phenomenon. 

\begin{figure}[ht]
\begin{center}
\includegraphics[width=0.6\linewidth]{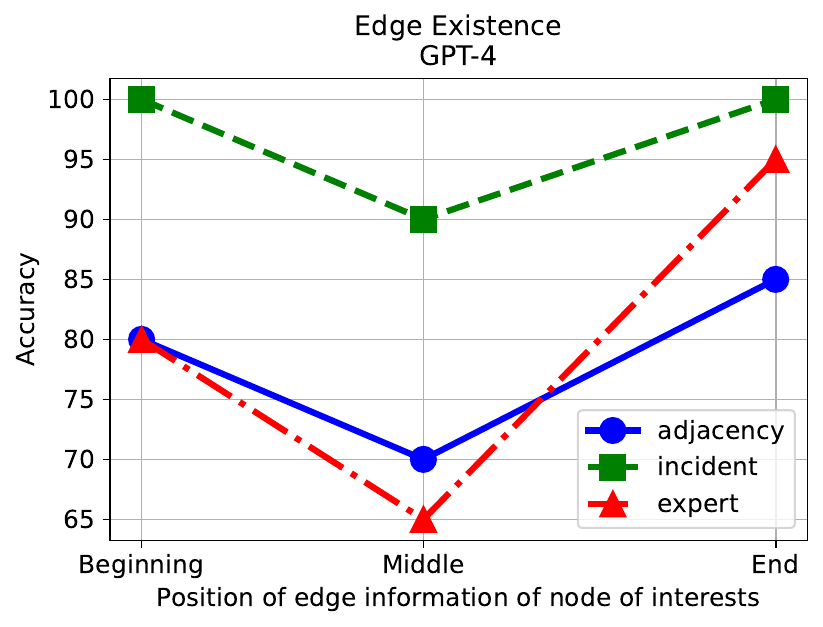}
\end{center}
\caption{The effect of the position of the relevant information on the edge existence task.}
\label{edge_node_fig}
\end{figure}

Figure~\ref{edge_node_fig} shows that all encodings can cause the LLM to lose the information in the middle of the prompt. The experiment results are averaged over twenty randomly generated graph where from each graph we randomly select two nodes and form the edge existence prompt as described in previous section. 

The best performance occurs when the relevant information is either at the beginning or the end of the entire subgraph structure.
Even for the incident encoding which has the best performance among all encodings, the LLM still has the worst performance when the answer is located in the middle of the prompt.

\section{Lost-in-Distance}
Tasks such as the edge existence require LLMs to perform needle-in-a-haystack retrieval, which, as previously shown, suffers from the lost-in-the-middle phenomenon in long contexts. However, in many tasks, the model not only needs to look up relevant information in the context but also requires to perform cross-referencing between retrieved information. For example, tasks like the common connection require the model to retrieve connections that \emph{jointly} appear in both subgraphs.

We hypothesize that for tasks requiring cross-referenced retrieval, the model's performance is also impacted by the distance between relevant pieces of information, a phenomenon we term lost-in-distance. Specifically, for these tasks, the model's performance is influenced by two compounding phenomena: lost-in-the-middle when retrieving relevant information and lost-in-distance when performing a join between retrieved information.

To explore this, we define $G(p)$ as the model's performance when the relevant information is at position $p$. Similarly, we define $F(p_1,p_2)$ as the model's performance when the relevant information is at positions $(p_1,p_2)$. The value of $F(p_1,p_2)$ is estimated based on the accuracy of model in a complex task that requires cross-referencing. We hypothesize that $F$ and $G$ have the following relationship:

\begin{equation}\label{eq:lost_in_distance_formula}
    F(p_1,p_2) = \gamma\, G(p_1) \, G(p_2) \, H(d),
\end{equation}

where $d=|p_2-p_1|$ represents the distance between relevant information in the prompt and $H(d)$ represents the effect of lost-in-distance.

In the experimental section, by studying LLM performance on common connection and similarity tasks, we first demonstrate that lost-in-the-middle alone cannot explain the model's performance degradation in solving tasks that require joint reasoning across multiple subgraphs, and that it is affected by another factor, lost-in-distance. Then, by leveraging the experimental results, we measure the goodness of fit for \Eqref{eq:lost_in_distance_formula} in Section~\ref{s:goodness_of_fite}.

\section{Experimentation}
In our initial experiments, we focused on the common connection task. This task requires the model to determine the number of common connections between two nodes by joining information across two subgraphs. Our results demonstrate that the models' performance degrades as the distance between the relevant pieces of information in the two subgraphs increases. Specifically, when the information about each node's connections is placed further apart in the context, the models struggle to effectively retrieve and integrate this information to compute the correct number of common connections.

We then investigated how the lost-in-distance impacts tasks that require multiple cross-referencing steps, such as the similarity task. In the similarity task, the model needs to first identify the common connections between each of the two nodes and a reference node, and then compare these sets to determine the degree of similarity. Our findings reveal that performance degradation is even more pronounced in this case, as the task requires the model to perform multiple join operations over dispersed pieces of information within the context.

\subsection{Experimental Setup}
Leveraging in-context learning \citep{dong2022survey, wei2023larger}, we conducted experiments using both closed-source models (GPT-4) and open-source models (Llama-3-8B-Instruct and Llama-3-70B-Instruct). For all models, we set the decoding temperature to zero to ensure the generation of deterministic answers. In each sample, we randomly selected two or three nodes as the nodes of interest for the common connection and similarity tasks, respectively. We performed experiments on We conducted experiments on one hundred thousand randomly generated graphs to derive statistically significant insights into LLM behavior. The experimental results were then averaged across multiple samples.

\subsection{Common Connection Task}
In this section, we demonstrate the effect of increased distance on solving the common connection task. To create an input prompt for this task and to control the relative distance of relevant information (common neighbors), we use the following methodology:

\begin{enumerate} 
    \item Sample two nodes from a given graph and extract their corresponding subgraphs. 
    \item Within each subgraph, group the common connections. 
    \item Within each subgraph, position the common connections at the beginning, middle, or end of the textual encoding of the subgraph.
\end{enumerate}

\begin{figure}[ht]
\begin{center}
\includegraphics[width=\linewidth]{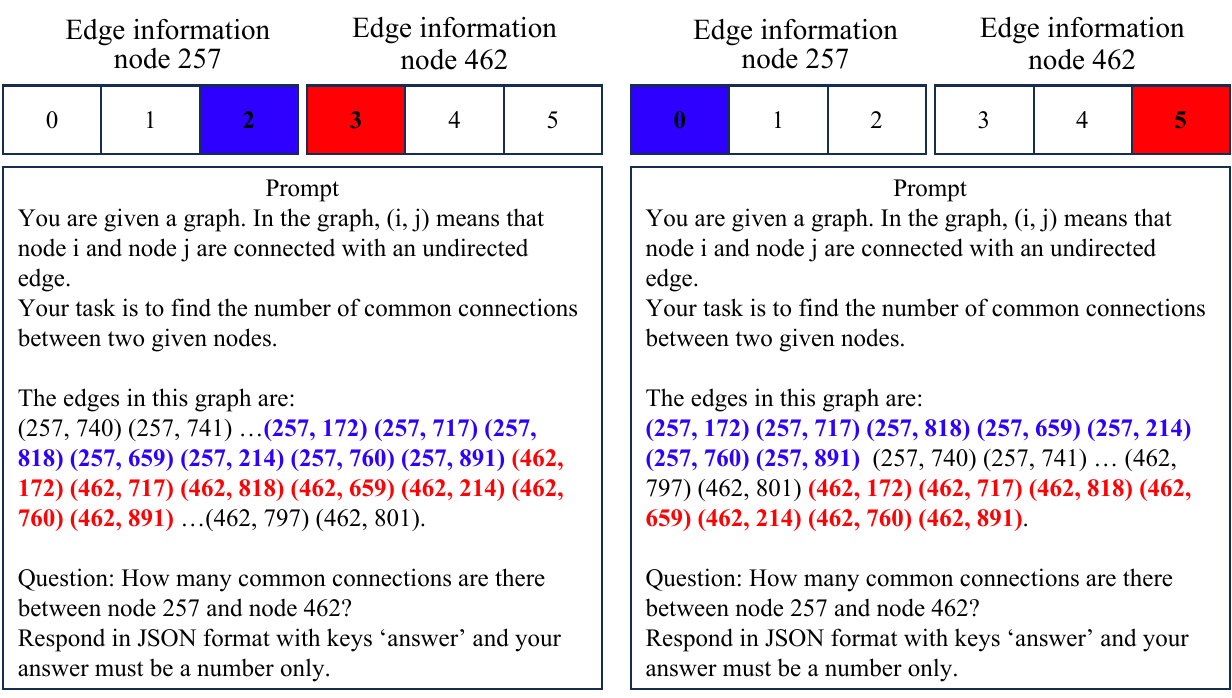}
\end{center}
\caption{An example illustrating the placement of relevant information, highlighted in blue and red, at different positions using the adjacency encoding function for the common connection task. Relevant information is grouped at positions $0$, $1$, or $2$ within the first node's (node $257$) subgraph and at positions $3$, $4$, or $5$ within the second node's (node $462$) subgraph. The left plot depicts the smallest distance between relevant information, while the right plot shows the largest distance.}
\label{common_connection_extreme_example_fig}
\end{figure}

The above recipe, specifically the grouping of relevant information into three positions—beginning, middle, and end (as illustrated in Figure~\ref{common_connection_extreme_example_fig} for adjacency encoding)—enables us to control the relative distance between common connections within the prompt. This allows us to investigate the effects on the model's performance when the relative distance is small, medium, or large. We denote the positions of relevant information within the first and second subgraphs as $(p_1, p_2)$, where $p_1 \in \{0,1,2\}$ and $p_2 \in \{3,4,5\}$, respectively, for the sake of brevity.

\subsubsection{Lost-in-Distance in Common Connection Task}
\begin{figure}[t]
    \centering
    \begin{subfigure}{0.33\linewidth}
        \centering
        \includegraphics[width=\linewidth]{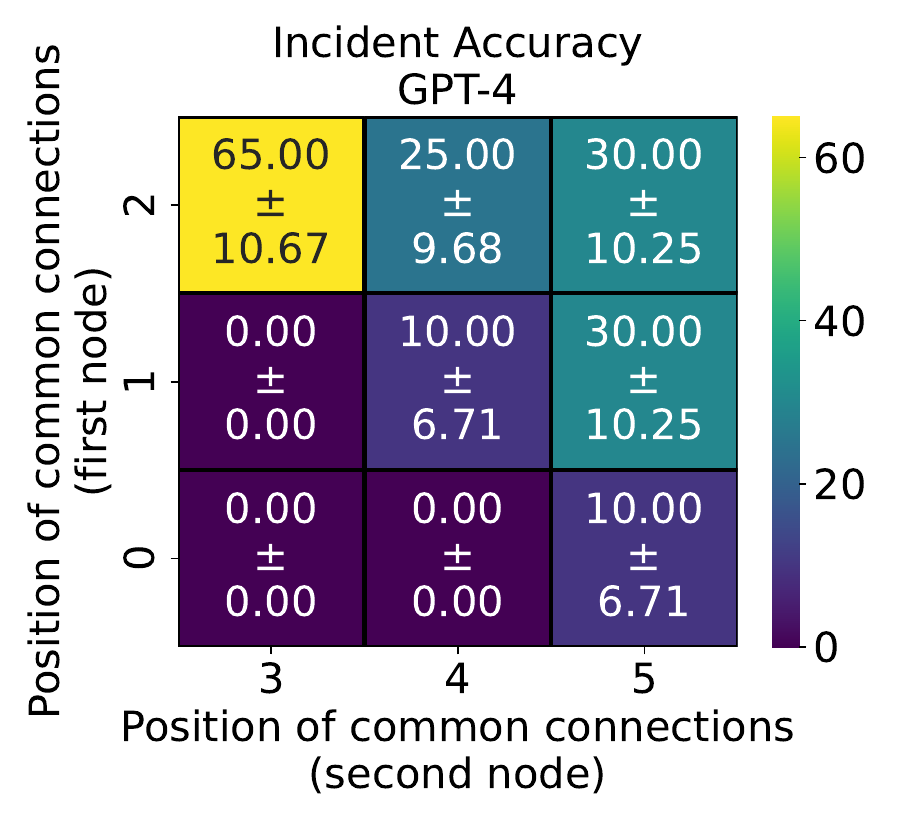}
    \end{subfigure}
    \hspace{-0.02\linewidth}
    \begin{subfigure}{0.33\linewidth}
        \centering
        \includegraphics[width=\linewidth]{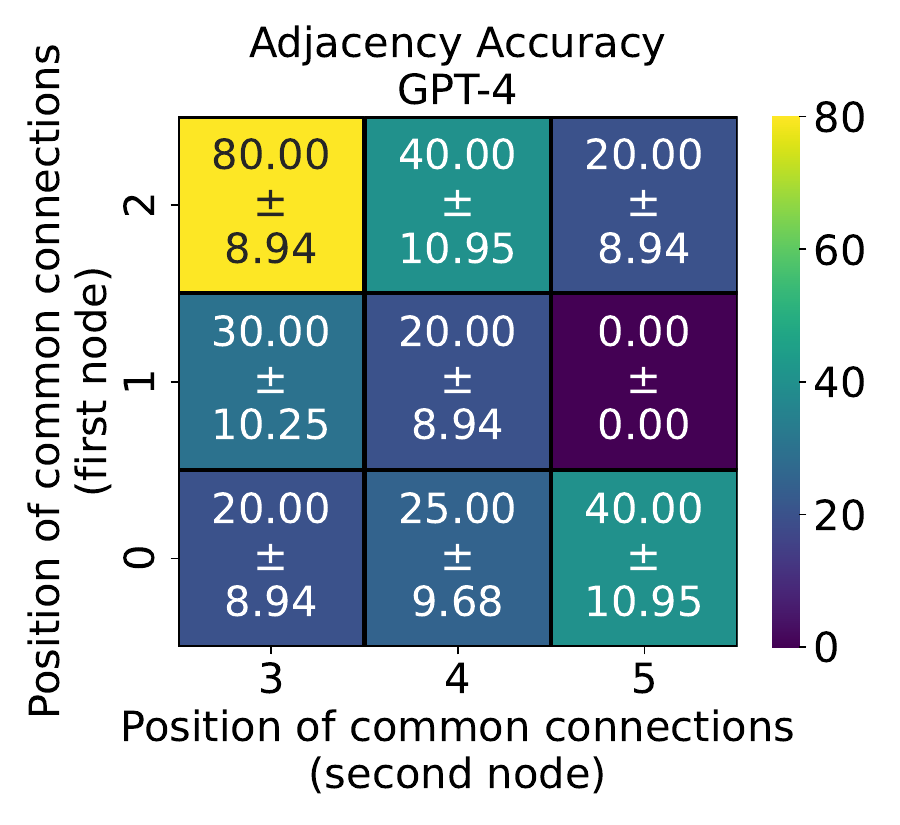}
    \end{subfigure}
    \hspace{-0.02\linewidth}
    \begin{subfigure}{0.33\linewidth}
        \centering
        \includegraphics[width=\linewidth]{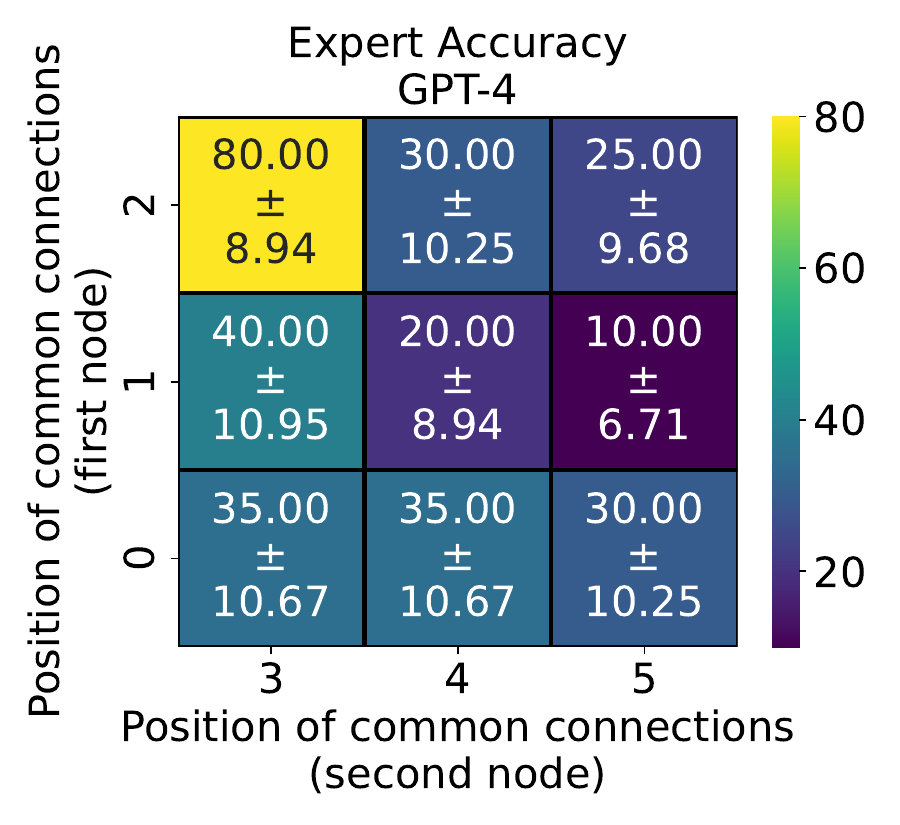}
    \end{subfigure}
    \caption{The effect of lost-in-distance on the common connection task. The number in each block is accuracy $\pm$ standard deviation.}
    \label{common_connection_extreme_GPT_4_fig}
\end{figure}

The results presented in Figure~\ref{common_connection_extreme_GPT_4_fig} illustrate the impact of varying the positions of common edges within each subgraph (following the methodology outlined in the previous section) on the model's performance in the common connection task. Unlike the edge existence task, the model's performance is influenced not only by the lost-in-the-middle phenomenon but also by the relative distance between common connections.

With the position of relevant information fixed in one subgraph, we observe that the model's performance degrades when the other subgraph is positioned closer to the middle of the prompt, influenced by the lost-in-the-middle phenomenon. For example, in adjacency encodings (Figure~\ref{common_connection_extreme_GPT_4_fig}, middle plot), when the first node's common connection is at position $0$ (the beginning of the prompt), the model's performance deteriorates from 40\% to 20\% as the second node's common connection shifts from position $5$ (the end) to position $3$ (the middle). However, in contrast to the lost-in-the-middle phenomenon, Figure~\ref{common_connection_extreme_GPT_4_fig} demonstrates that across all three graph encodings, the model achieves optimal performance when relevant information is centrally located, with minimal distance between components at positions $(2,3)$. This illustrates the effect of lost-in-distance. Furthermore, when the first node's common connection is at position $2$, the model's accuracy drops by up to 50\% as the second node's common connection shifts from position $3$ (the middle of the prompt) to position $5$ (the end of the prompt), thereby increasing the distance between relevant information. These observations confirm that the lost-in-distance phenomenon and the lost-in-the-middle effect have independent, compounding effects on model performance, as hypothesized in~\Eqref{eq:lost_in_distance_formula}.

\subsection{Similarity Task}\label{ss:lost_in_distance_for_similarity}
Solving the similarity between three nodes $v_i$, $v_j$, and $v_k$ requires the model to perform two common connection tasks, $|\mathcal{N}(v_{i}) \cap \mathcal{N}(v_{j})|$ and $|\mathcal{N}(v_{j}) \cap \mathcal{N}(v_{k})|$, and subsequently compare the results. As a result, the model needs to execute two cross-referencing operations between the subgraphs: one between $\mathcal{G}_{v_i}$ and $\mathcal{G}_{v_j}$, and the other between $\mathcal{G}_{v_j}$ and $\mathcal{G}_{v_k}$. Therefore, as we will demonstrate in this section, solving the similarity task inherently suffers from the lost-in-distance phenomenon.

To measure the effect of the lost-in-distance phenomenon, we select three nodes—$v_{i}$, $v_{j}$, and $v_{k}$—from a given graph and randomly shuffle the edges within each node's subgraph. We designate $v_{j}$ as the source node for the similarity task, $v_{i}$ as the first target node, and $v_{k}$ as the second target node. In all scenarios, to mitigate the influence of the lost-in-the-middle effect and highlight the effect of lost-in-distance, the textual encoding of the source node $v_{j}$'s subgraph ($\mathcal{G}_{v_j}$) is positioned at the center of the prompt, while the subgraphs of the other two nodes are placed one before and one after it.

\begin{figure}[b]
    \centering
    \begin{subfigure}{0.33\linewidth}
        \centering
        \includegraphics[width=\linewidth]{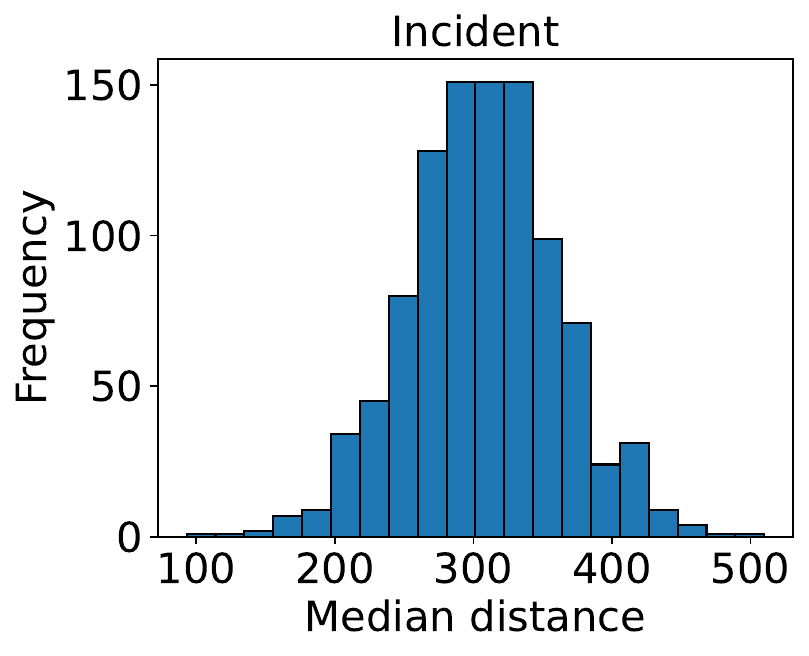}
    \end{subfigure}
    \hspace{-0.01\linewidth}
    \begin{subfigure}{0.33\linewidth}
        \centering
        \includegraphics[width=\linewidth]{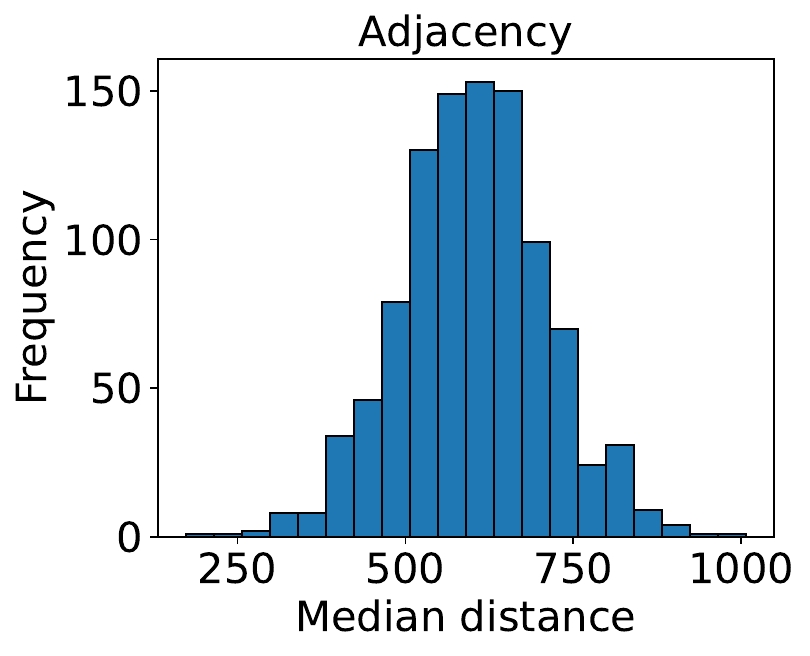}
    \end{subfigure}
    \hspace{-0.01\linewidth}
    \begin{subfigure}{0.33\linewidth}
        \centering
        \includegraphics[width=\linewidth]{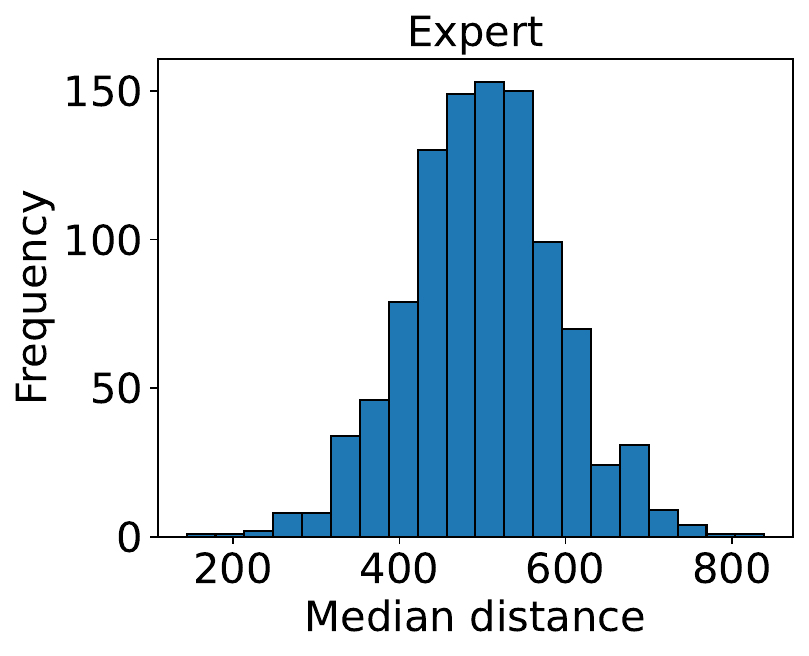}
    \end{subfigure}
    \caption{The distribution of median distance, in number of tokens, for three graph encoding functions.}
    \label{fig:distance_distribution_fig}
\end{figure}

We quantify the lost-in-distance effect by calculating the median distance,  measured in terms of tokens, between the common connections of the two subgraphs, specifically $|\mathcal{N}(v_{i}) \cap \mathcal{N}(v_{j})|$ and $|\mathcal{N}(v_{j}) \cap \mathcal{N}(v_{k})|$. The distance distribution is illustrated in Figure~\ref{fig:distance_distribution_fig} for three different graph encodings. We utilize the thresholds presented in Table~\ref{t:thresholds_distance_table} to categorize the distances into small, medium, and large groups. Furthermore, in order to make sure more uniform coverage, we employ rejection sampling to ensure that each distance group contains one hundred samples with balanced responses.

\begin{table}[t]
\centering
\caption{Thresholds of each distance group for three graph encoding functions where distance is measured in number of tokens.}
\label{t:thresholds_distance_table}
\resizebox{0.7\linewidth}{!}{%
\begin{tabular}{cccc}
\hline
Graph Encoding & Small Distance & Medium Distance & Large Distance \\ \hline
Incident       & $\le 219$      & $219 \sim 399$  & $> 399$        \\
Adjacency      & $\le 425$      & $425 \sim 785$  & $> 785$        \\
Expert         & $\le 354$      & $354 \sim 654$  & $> 654$        \\ \hline
\end{tabular}%
}
\end{table}

To eliminate potential biases, for the three subgraphs $\mathcal{G}_{v_{i}}$, $\mathcal{G}_{v_{j}}$, and $\mathcal{G}_{v_{k}}$, where $v_j$ is the source node for similarity, we generate questions randomly chosen from the following two templates:
\begin{itemize}
    \item \textit{Is the number of common connections between \textbf{node $\boldsymbol{v_{j}}$ and node $\boldsymbol{v_{k}}$} greater than the number of common connections between \textbf{node $\boldsymbol{v_{i}}$ and node $\boldsymbol{v_{j}}$}?}
    \item \textit{Is the number of common connections between \textbf{node $\boldsymbol{v_{i}}$ and node $\boldsymbol{v_{j}}$} greater than the number of common connections between \textbf{node $\boldsymbol{v_{j}}$ and node $\boldsymbol{v_{k}}$}?} 
\end{itemize}
An example of the prompt for the similarity task is presented in Appendix~\ref{s:appendix:similarity_prompt_example}. Since the similarity task inherently involves solving two common connection tasks and a comparison task, we adopt zero-shot step-by-step reasoning prompting \citep{kojima2022large} to guide the model in solving the task step by step, thereby obtaining the most accurate answers. Detailed descriptions of the prompts and further analysis of the LLMs' failure rates in following zero-shot reasoning instructions, along with examples of answer degeneration, are provided in Appendix~\ref{ss:appendix:answer_degeneration}. GPT-4 and Llama-3-70B exhibit low failure rates, while Llama-3-8B demonstrates a high failure rate across all graph encodings, including a 69\% failure rate in expert encodings.

\subsubsection{Lost-in-Distance in Similarity Task}
For brevity, we present the results of one encoding for each model in Figure~\ref{similarity_median_fig}, with all results summarized in Appendix~\ref{s:appendix:all_results}. Our findings indicate that when both distances are minimal, GPT-4 and Llama-3-70B-Instruct exhibit the best performance. Llama-3-8B-Instruct, which has a high failure rate in following instructions as described in Appendix~\ref{ss:appendix:answer_degeneration}, demonstrates the second-best performance, though it is not significantly different from the top performers.

\begin{figure}[t]
    \centering
    \begin{subfigure}{0.33\linewidth}
        \centering
        \includegraphics[width=\linewidth]{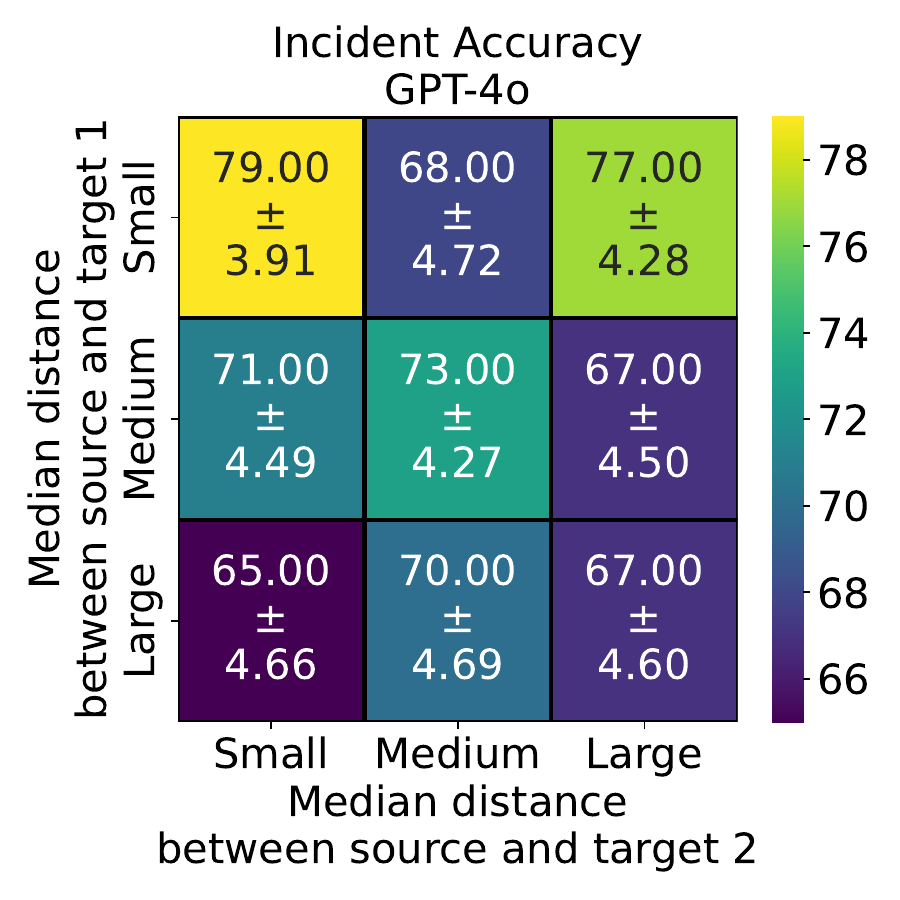}
    \end{subfigure}
    \hspace{-0.02\linewidth}
    \begin{subfigure}{0.33\linewidth}
        \centering
        \includegraphics[width=\linewidth]{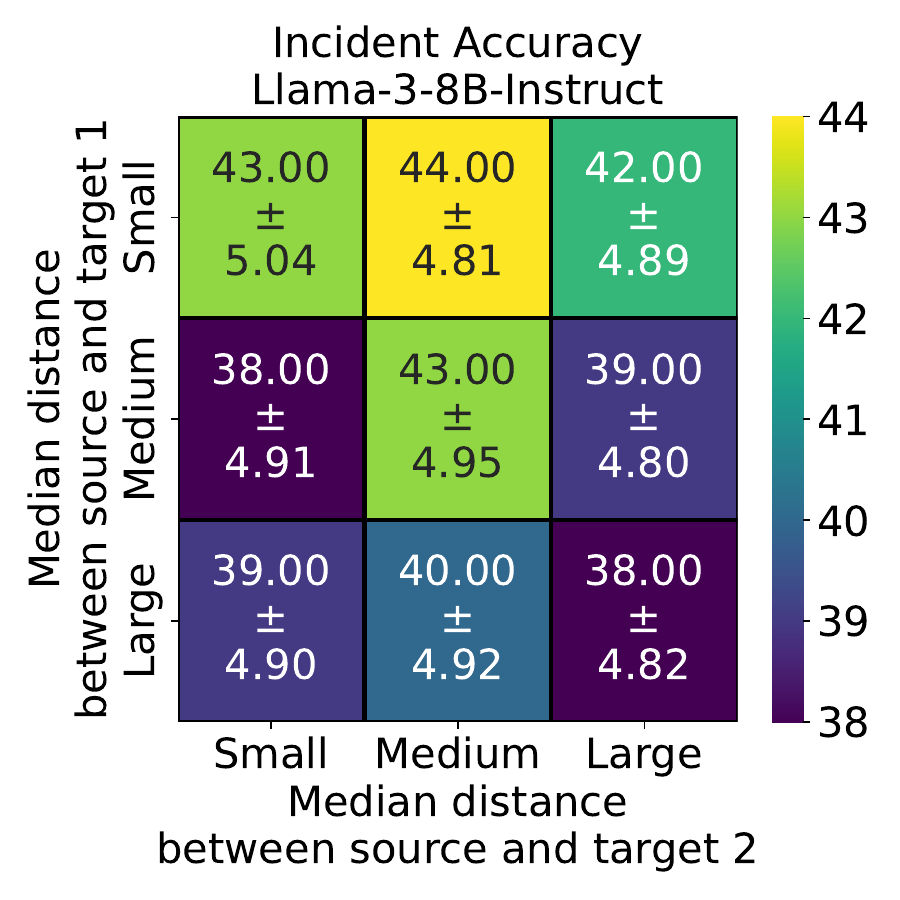}
    \end{subfigure}
    \hspace{-0.02\linewidth}
    \begin{subfigure}{0.33\linewidth}
        \centering
        \includegraphics[width=\linewidth]{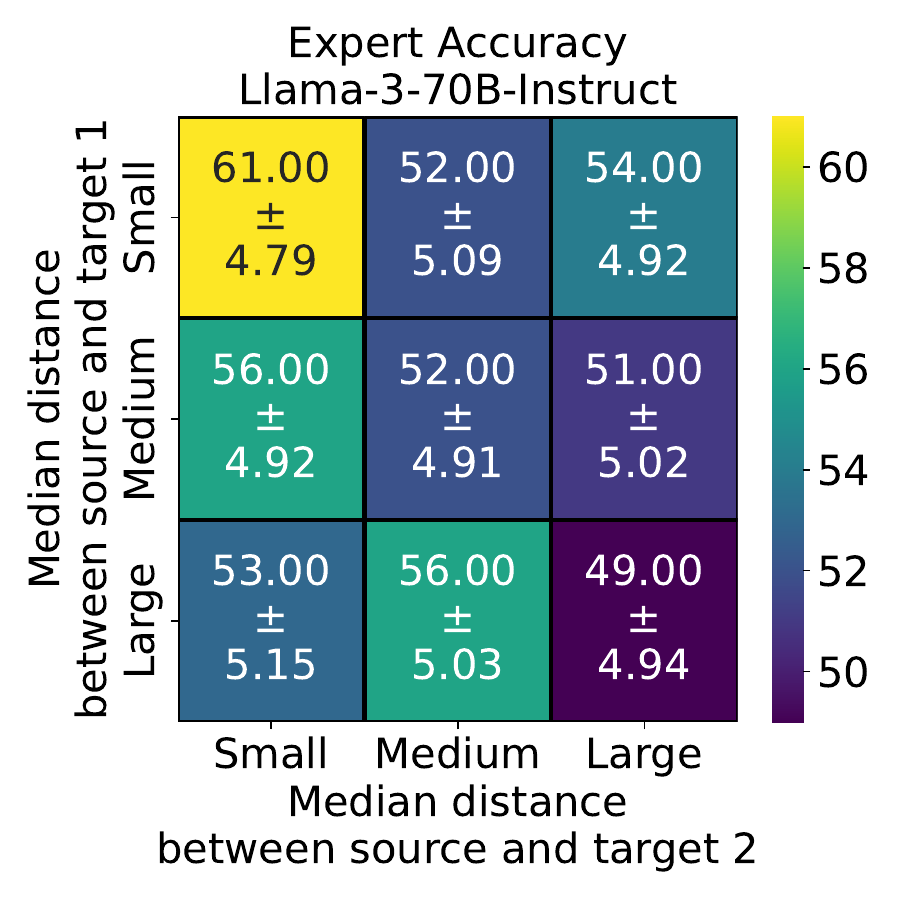}
    \end{subfigure}
    \caption{The effect of lost-in-distance on the similarity task is illustrated. As the distance between target node 1 and target node 2 increases, the model's performance degrades accordingly. The numbers in each block represent the accuracy $\pm$ standard deviation obtained through bootstrap sampling.}
    \label{similarity_median_fig}
\end{figure}

Specifically, performance at the largest distances is significantly worse compared to that at the smallest distances. As the distances increase (i.e., along the diagonal elements), the performance of all models deteriorates. In Llama-3-70B, we observe a 12\% drop in model accuracy when the distance between common connections for both $|\mathcal{N}(v_{i}) \cap \mathcal{N}(v_{j})|$ and $|\mathcal{N}(v_{j}) \cap \mathcal{N}(v_{k})|$ increases, shifting from the $(\text{Small}, \text{Small})$ index to the $(\text{Large}, \text{Large})$ index in the heatmap plot. These results highlight that the lost-in-distance phenomenon adversely affects model performance in similarity tasks.

\section{Goodness of Fit for Lost-in-Distance}\label{s:goodness_of_fite}
In this section, we employ the~\Eqref{eq:lost_in_distance_formula} function to capture the effects of the lost-in-distance phenomenon and to separate its impact from that of the lost-in-the-middle effect. To evaluate the goodness of fit for the lost-in-distance function defined in~\Eqref{eq:lost_in_distance_formula}, we compare it to a simpler function that accounts solely for the lost-in-the-middle effect as follows:
\begin{equation}
\label{simple_model}
    \E[F(p_1,p_2)|G(p_1),G(p_2)] = \gamma G(p_1) G(p_2),
\end{equation}
\begin{equation}
\label{complex_model}
    \E[F(p_1,p_2)|\gamma, G(p_1), G(p_2)] = \gamma G(p_1) G(p_2) H(|p_2-p_1|),
\end{equation}
where $H(|p_2-p_1|))$ is the effect of lost-in-distance $d=|p_2-p_1|$.

To measure the goodness of fit we leverage results and output of common connection experiments but the result and findings here are extendable to similarity task as well. We randomly split samples into training and test sets of equal size. Using the training set, we first estimate $\hat{G}(\cdot)$ using interpolation based on the accuracy observed in the edge existence task. Then, we estimate $\gamma$ by regressing $F(p_1,p_2)$ onto $\hat{G}(p_1)\hat{G}(p_2)$. Finally, given the estimated $\hat{\gamma}$ and $\hat{G}(\cdot)$, we estimate $H(\cdot)$ using
\begin{equation}
    \hat{H}(d) = \frac{1}{|\gD_{d}|}\sum_{(p_1,p_2)\in \gD_{d}} \frac{F(p_1,p_2)}{\hat{\gamma}\hat{G}(p_1)\hat{G}(p_2)},
\end{equation}
where $\gD_{d} = \{(p_1,p_2)||p_2-p_1|=d\}$.

We evaluate the goodness of fit for both functions using the root mean squared error (RMSE) between the predicted and observed accuracy in the test set.

\begin{figure}[ht]
    \centering
    \begin{subfigure}{\linewidth}
        \centering
        \includegraphics[width=\linewidth]{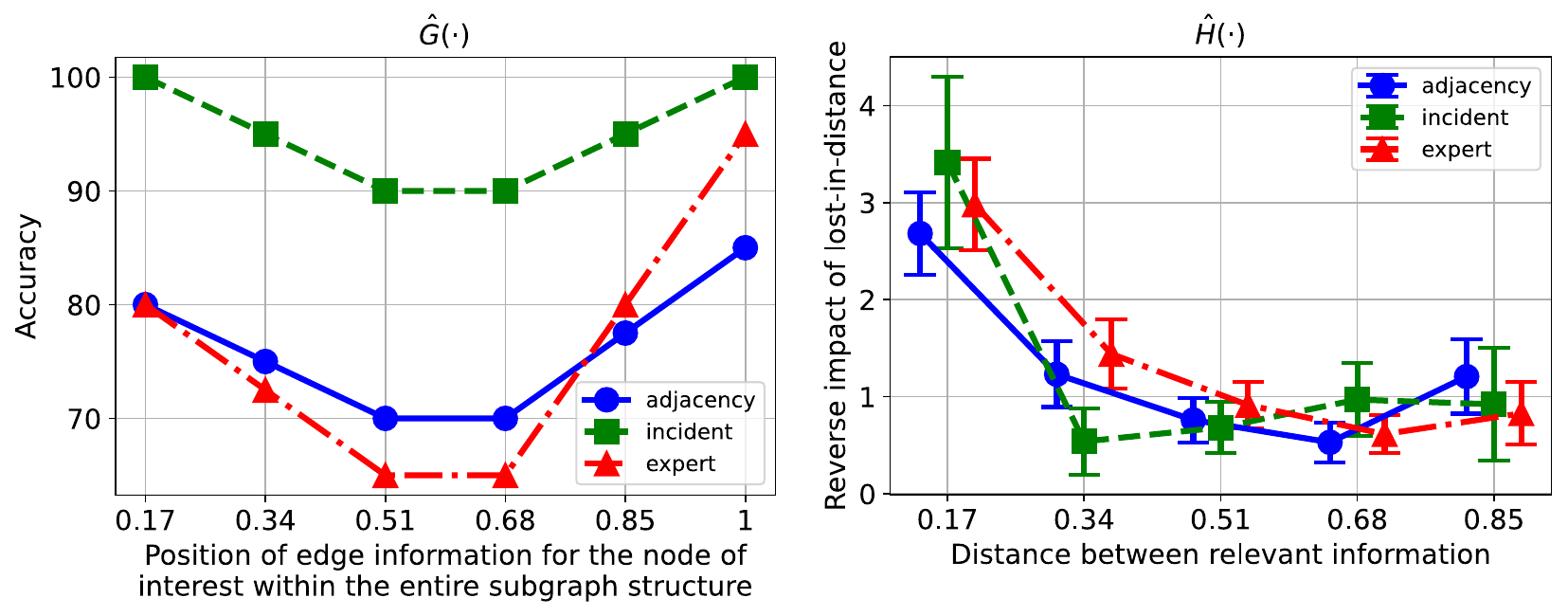}
    \end{subfigure}
    \caption{The left plot is $\hat{G}(\cdot)$ and the right plot is $\hat{H}(\cdot)$. To better visualize the error bars in $\hat{H}(\cdot)$ estimation, we slightly shift the adjacency encoding to the left and the expert encoding to the right. Numbers in x-axis are normalized to the prompt length.}
    \label{goodness_of_fit_fig}
\end{figure}

Table~\ref{goodness_of_fit_table} shows that~\Eqref{eq:lost_in_distance_formula} function, which includes the lost-in-distance effect, has a smaller RMSE compared to the lost-in-the-middle only function. 
Moreover, $\hat{H}(\cdot)$ in Figure~\ref{goodness_of_fit_fig} indicates that smaller distance results in better performance, up to 3x, after accounting for the lost-in-the-middle effect.

\begin{table}[t]
\centering
\caption{RMSE for lost-in-the-middle only and~\Eqref{eq:lost_in_distance_formula} functions with three encodings.}
\label{goodness_of_fit_table}
\resizebox{0.6\linewidth}{!}{%
\begin{tabular}{cccc}
\hline
Encoding & Model & RMSE (train) & RMSE (test) \\ \hline
\multirow{2}{*}{Incident}  & Lost-in-the-middle only  & 22.42 & 27.09 \\
                           & \Eqref{eq:lost_in_distance_formula} & 10.04 & 14.50 \\
\multirow{2}{*}{Adjacency} & Lost-in-the-middle only  & 21.43 & 22.36 \\
                           & \Eqref{eq:lost_in_distance_formula} & 15.16 & 13.61 \\
\multirow{2}{*}{Expert}    & Lost-in-the-middle only  & 24.50 & 26.69 \\
                           & \Eqref{eq:lost_in_distance_formula} & 12.84 & 17.42 \\ \hline
\end{tabular}%
}
\end{table}

\section{Conclusion}
This study introduces the "lost-in-distance" phenomenon, demonstrating how the performance of LLMs in graph-related tasks is impacted by the relative positioning of relevant information in the input context. Our experiments reveal that as the distance between crucial data points increases, the accuracy of publicly available open and proprietary LLMs significantly deteriorates in solving graph tasks that require cross-referencing. This phenomenon compounds with the known "lost-in-the-middle" effect, leading to notable declines in task performance when multiple subgraphs must be cross-referenced, such as in common connection and similarity tasks.

Our findings, which we coin "lost-in-distance," highlight a key limitation of current LLMs in solving graph tasks, particularly those requiring complex reasoning over dispersed data. This study paves the way for future research focused on mitigating these issues, such as developing advanced graph encoding techniques. Addressing the challenges posed by lost-in-distance will be critical for enhancing LLM applications in domains like recommendation systems, molecular design, and multi-hop question answering, and will contribute to the broader field of LLMs in graph problem-solving.

\bibliography{iclr2025_conference}
\bibliographystyle{iclr2025_conference}

\newpage
\appendix
\section{Analysis} \label{s:appendix}

\subsection{Context Length}
Table~\ref{context_length_table} summarizes the average context length (i.e., the number of tokens) for each task and each graph encoding. We use the tokenizer of Llama-3 to calculate the context length for Llama-3-8B-Instruct and Llama-3-70B-Instruct and use the tiktoken library \citep{openai_tiktoken} to calculate the context length for GPT-4 and GPT-4o. The incident encoding produces the shortest context length, while the adjacency encoding results in the longest context length. 

\begin{table}[htbp]
\centering
\caption{Input and output context length in each encoding and each task.}
\label{context_length_table}
\resizebox{\textwidth}{!}{%
\begin{tabular}{cccccc}
\hline
\multirow{2}{*}{Graph Task} & \multirow{2}{*}{Graph Encoding} & \multirow{2}{*}{Average Input Length} & \multicolumn{3}{c}{Average Output Length} \\ \cline{4-6} 
            & & & GPT-4/4o & Llama-3-8B-Instruct & Llama-3-70B-Instruct \\ \hline
\multirow{3}{*}{Edge Existence}  & Incident  & 3409.50 & 13 & 8 & 7  \\
                                 & Adjacency & 6598.10 & 13 & 8 & 7   \\
                                 & Expert    & 5514.75 & 13 & 8 & 7    \\
\multirow{3}{*}{Common Connection} & Incident & 662.55 & 13 & 8 & 7    \\
                                   & Adjacency & 1261.10 & 13 & 8 & 7  \\
                                   & Expert    & 1068.25 & 13 & 8 & 7   \\
\multirow{3}{*}{Similarity} & Incident  & 1263.37 & 153.46 & 695.31 & 91.76 \\
                            & Adjacency & 2164.75 & 142.32 & 1126.30 & 106.86 \\
                            & Expert    & 1869.29 & 140.34 & 1453.97 & 97.73   \\ \hline
\end{tabular}%
}
\end{table}

\subsection{Answer degeneration}\label{ss:appendix:answer_degeneration}
\begin{figure}[ht]
\begin{center}
\includegraphics[width=\linewidth]{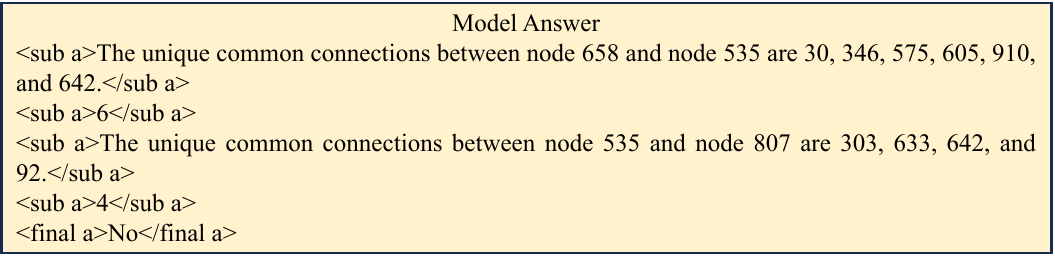}
\end{center}
\caption{Example where the answer is self-contradictory. The question in this example is that ``\textit{is the number of common connections between \textbf{node 658 and node 535} greater than the number of common connections between \textbf{node 535 and node 807}?}".}
\label{contradicted_answer_example_fig}
\end{figure}

LLMs sometimes fail to follow instructions and generate responses that do not adhere to the expected output template. We classify these degenerate responses as incorrect answers, i.e., $y_{i} \neq \hat{y}_{i}$ for accuracy calculation in \Eqref{math:accuracy}. Llama-3-8B-Instruct is less likely to generate a final answer compared to GPT-4o and Llama-3-70B-Instruct, which explains why it has lowest accuracy in Figure~\ref{similarity_median_fig}. Table~\ref{t:degenerated_answer} summarizes the percentage of samples in which models fail to follow instructions. Generally, the most common patterns of degenerate answers are as follows:

\begin{itemize}
    \item \textbf{Repetition}: LLMs sometimes repeat the same context until they reach the maximum number of output tokens, thereby failing to generate a final answer.
    \item \textbf{Self-contradiction}: In zero-shot step-by-step reasoning prompting~\citep{kojima2022large}, LLMs are asked to answer the main question based on their responses to subquestions. However, we find that LLMs sometimes provide an incorrect final conclusion. For example, as shown in Figure~\ref{contradicted_answer_example_fig} where we ask the LLM ``\textit{is the number of common connections between \textbf{node 658 and node 535} greater than the number of common connections between \textbf{node 535 and node 807}?}", the LLM determines that the number of common connections between node $658$ and node $535$ is $6$, and between node $535$ and node $807$ is $4$, but the final answer is ``no" when it should be ``yes".
\end{itemize}

\begin{table}[t]
\centering
\caption{Percentage of samples where each model generates degenerated answer.}
\label{not_generate_answer_table}
\resizebox{0.8\linewidth}{!}{%
\begin{tabular}{cccc}
\hline
\multicolumn{1}{l}{\multirow{2}{*}{Graph Encoding}} & \multicolumn{3}{c}{Percentage of not generating the answer} \\ \cline{2-4} 
\multicolumn{1}{l}{} & GPT-4o & Llama-3-8B-Instruct & Llama-3-70B-Instruct   \\ \hline
Incident  & $0.78$\% & $18.78$\% & $0.11$\% \\
Adjacency & $4.11$\% & $61.11$\% & $0.22$\% \\
Expert    & $4.44$\% & $69.00$\% & $0.22$\% \\ \hline
\end{tabular}%
}
\end{table}\label{t:degenerated_answer}

\section{Common Connection: More result}\label{s:appendix:common_connection_more_result}
\begin{figure}[ht]
    \centering
    \begin{subfigure}{0.33\linewidth}
        \centering
        \includegraphics[width=\linewidth]{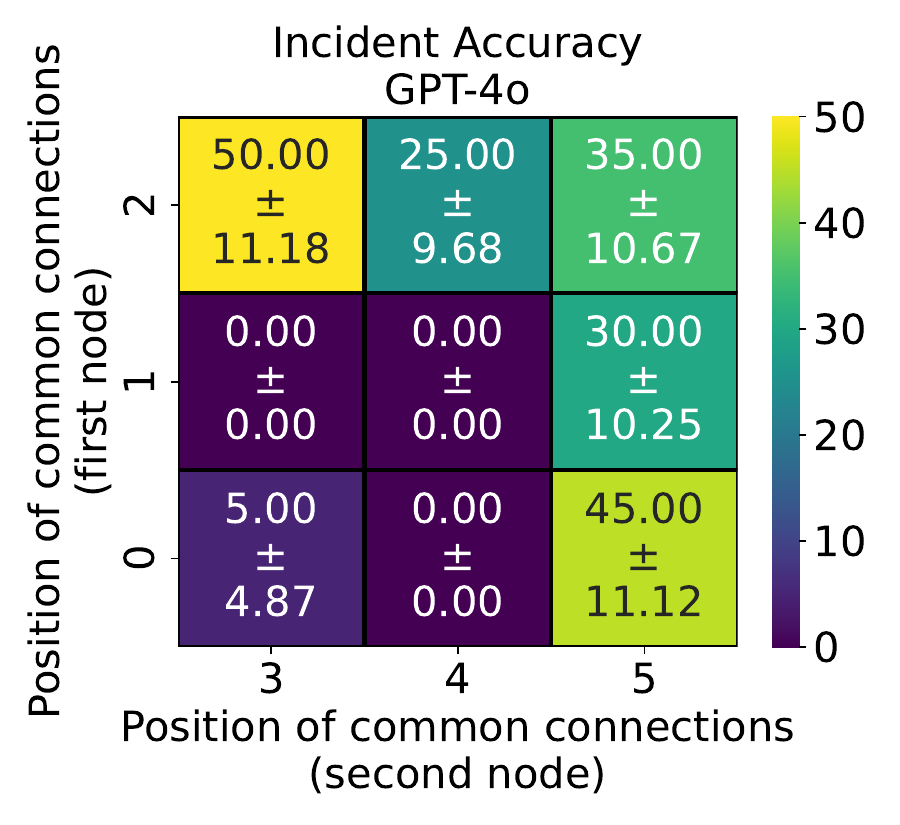}
    \end{subfigure}
    \hspace{-0.02\linewidth}
    \begin{subfigure}{0.33\linewidth}
        \centering
        \includegraphics[width=\linewidth]{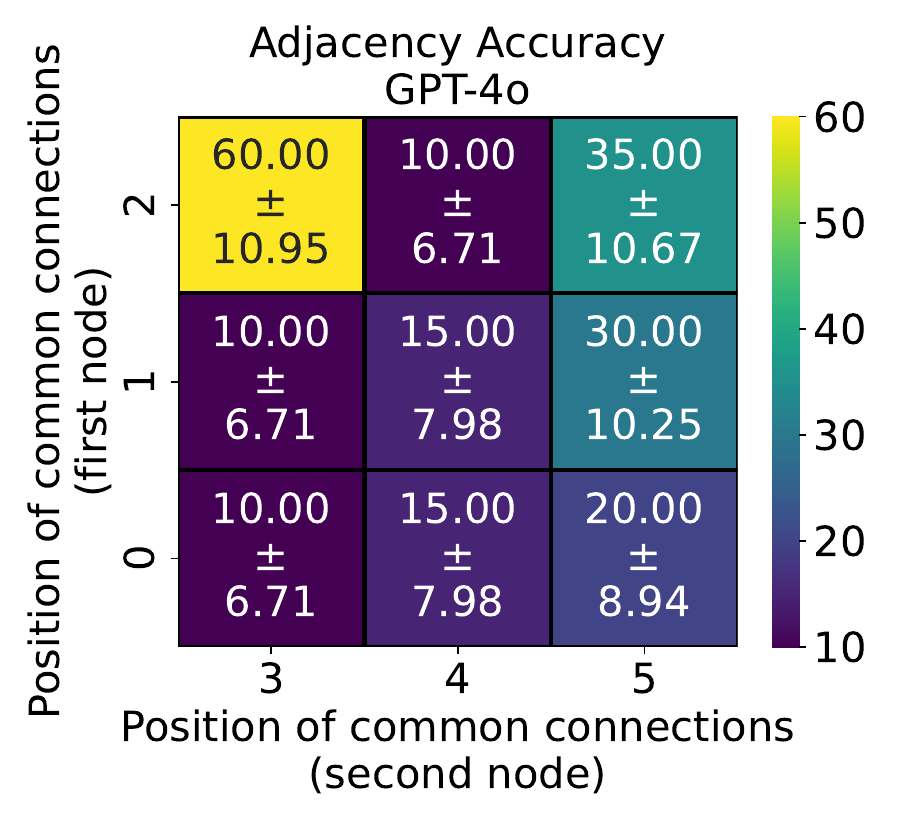}
    \end{subfigure}
    \hspace{-0.02\linewidth}
    \begin{subfigure}{0.33\linewidth}
        \centering
        \includegraphics[width=\linewidth]{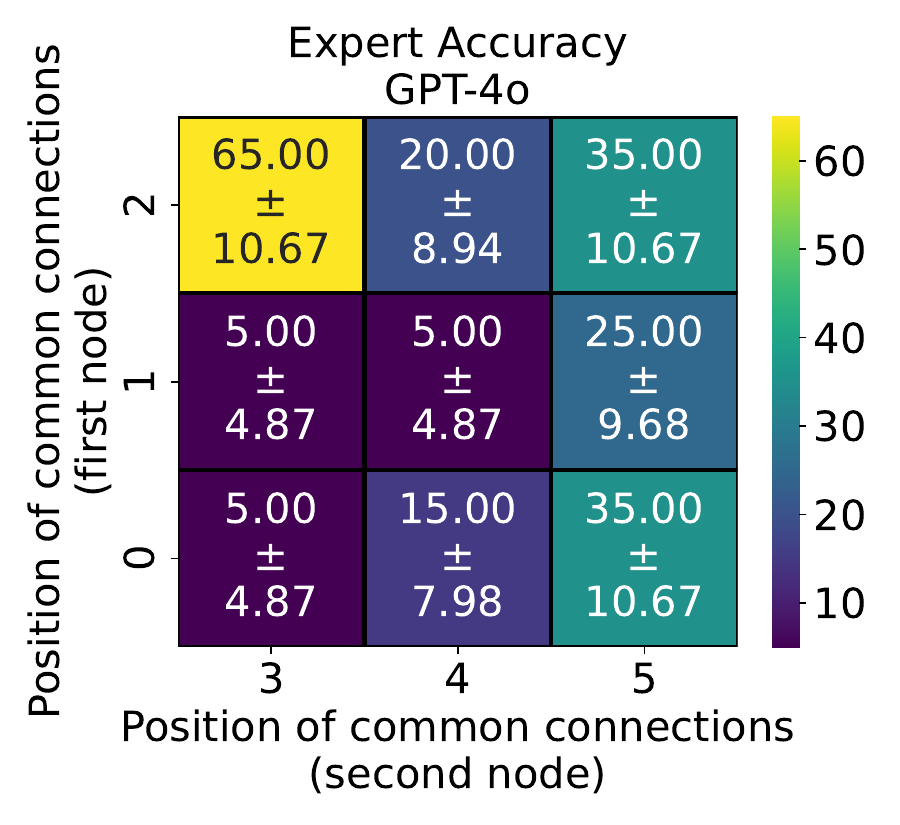}
    \end{subfigure}
    \caption{The effect of the position of the relevant information on the GPT-4o model solving the common connection task.}
    \label{fig:common_connection_extreme_GPT_4o_fig}
\end{figure}

Figure~\ref{fig:common_connection_extreme_GPT_4o_fig} illustrates the impact of the lost-in-distance phenomenon on the GPT-4o model \citep{openai_gpt4o} in solving the common connection task. Although the accuracy metrics slightly differ from those in Figure~\ref{common_connection_extreme_GPT_4_fig} for GPT-4, the pattern of the lost-in-distance effect remains consistent.

\section{Similarity Task}\label{s:appendix:all_res_similarity_0.1}
\subsection{Prompt Example}\label{s:appendix:similarity_prompt_example}
\begin{figure}[ht]
\begin{center}
\includegraphics[width=\linewidth]{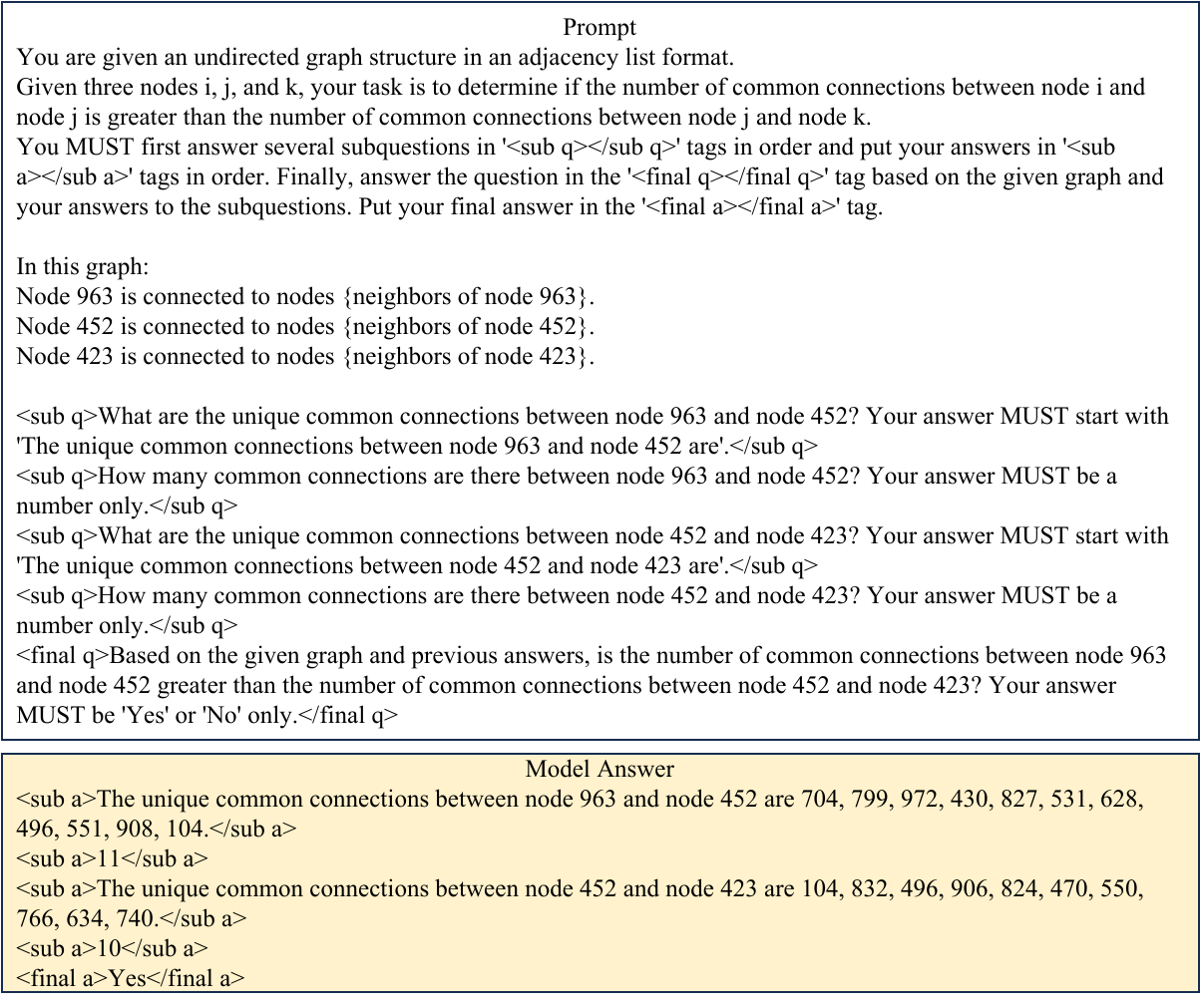}
\end{center}
\caption{Example of the zero-shot step-by-step reasoning prompting for the similarity task.}
\label{fig:CoT_example_fig}
\end{figure}

Figure~\ref{fig:CoT_example_fig} illustrates an example of the similarity task prompt, as described in Section~\ref{ss:lost_in_distance_for_similarity}, along with GPT-4o's answer for solving the similarity task using incident graph encoding.

\subsection{All results}\label{s:appendix:all_results}
\begin{figure}[ht]
    \centering
    \begin{subfigure}{0.33\linewidth}
        \centering
        \includegraphics[width=\linewidth]{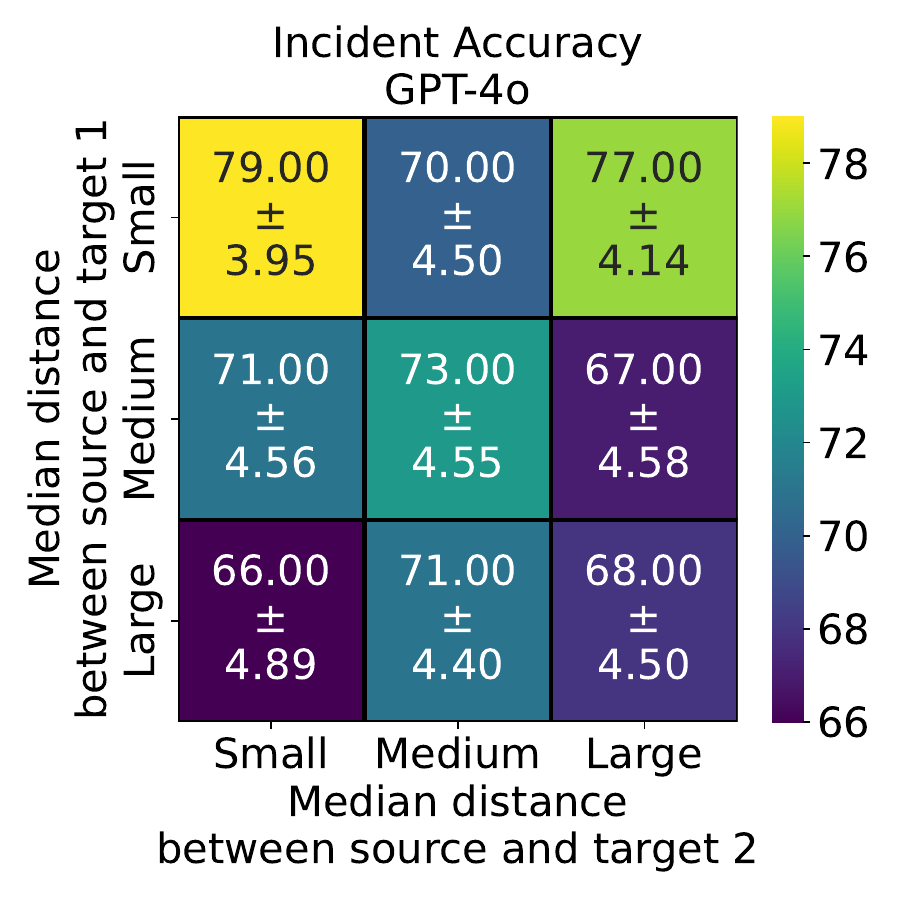}
    \end{subfigure}
    \hspace{-0.02\linewidth}
    \begin{subfigure}{0.33\linewidth}
        \centering
        \includegraphics[width=\linewidth]{./Figures/similarity_median_correct/GPT-4o-adjacency_correct.pdf}
    \end{subfigure}
    \hspace{-0.02\linewidth}
    \begin{subfigure}{0.33\linewidth}
        \centering
        \includegraphics[width=\linewidth]{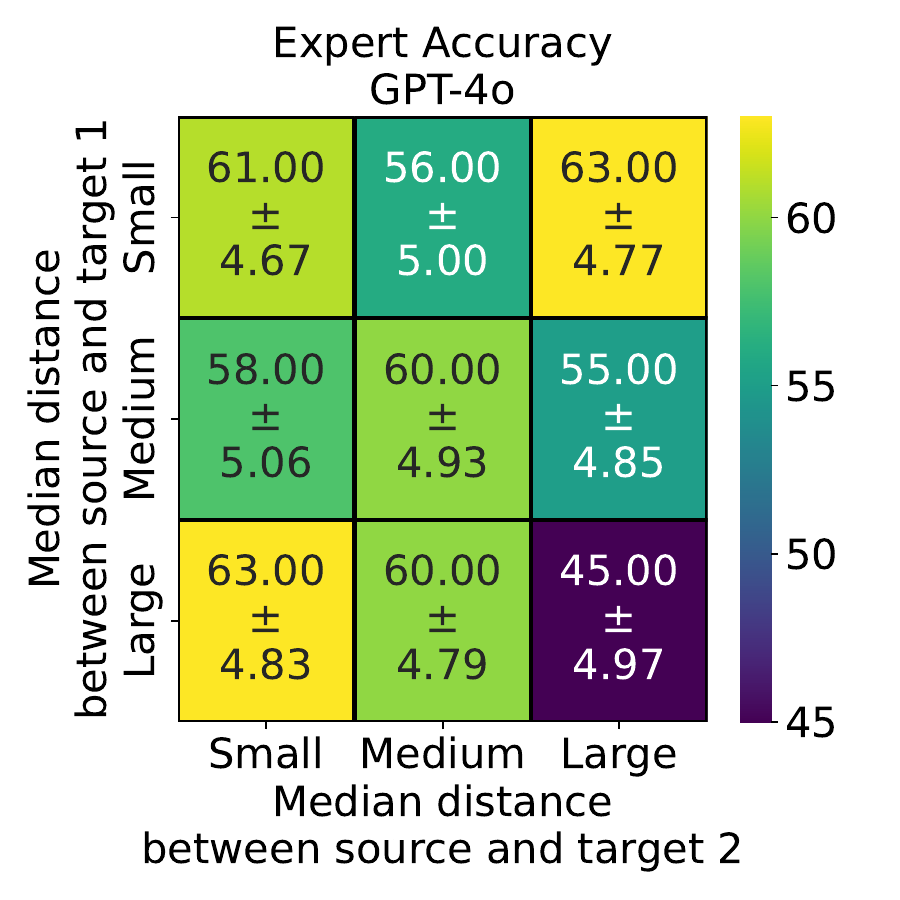}
    \end{subfigure}

        \begin{subfigure}{0.33\linewidth}
        \centering
        \includegraphics[width=\linewidth]{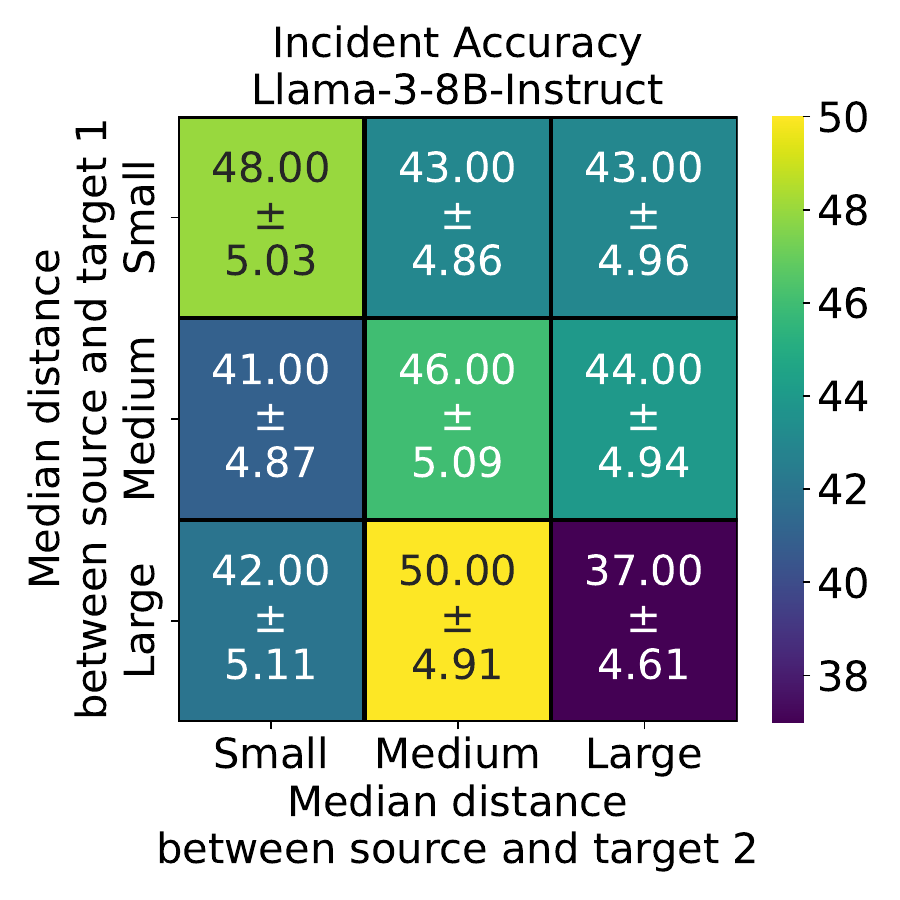}
    \end{subfigure}
    \hspace{-0.02\linewidth}
    \begin{subfigure}{0.33\linewidth}
        \centering
        \includegraphics[width=\linewidth]{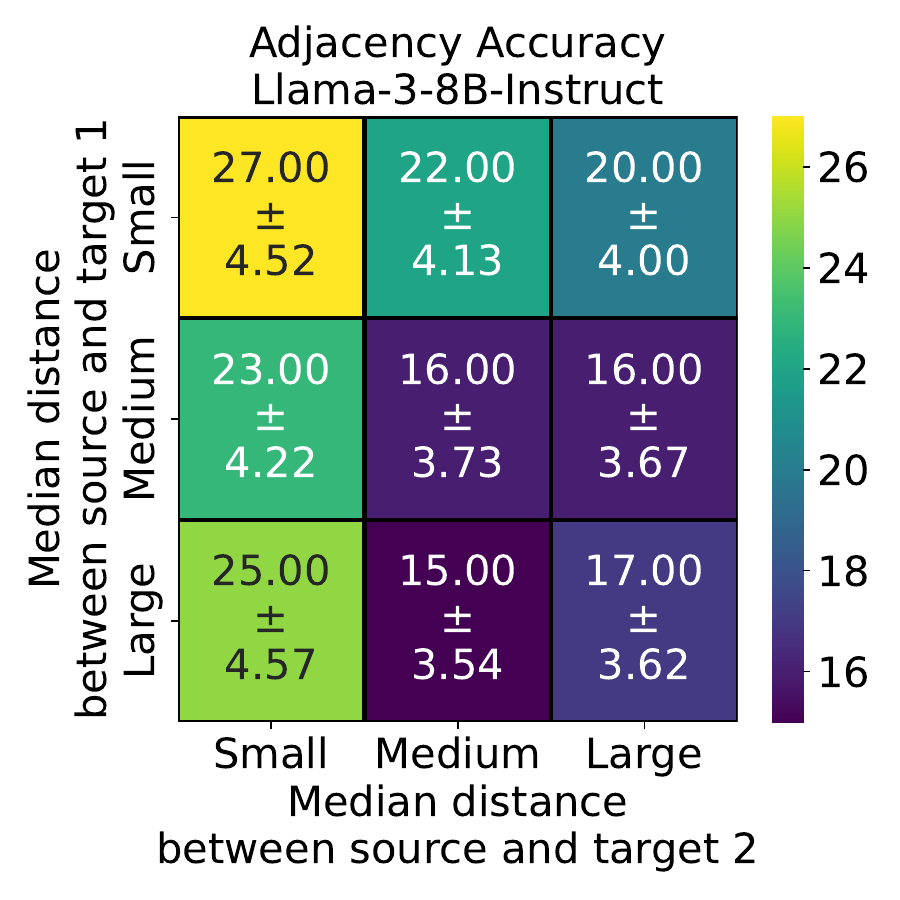}
    \end{subfigure}
    \hspace{-0.02\linewidth}
    \begin{subfigure}{0.33\linewidth}
        \centering
        \includegraphics[width=\linewidth]{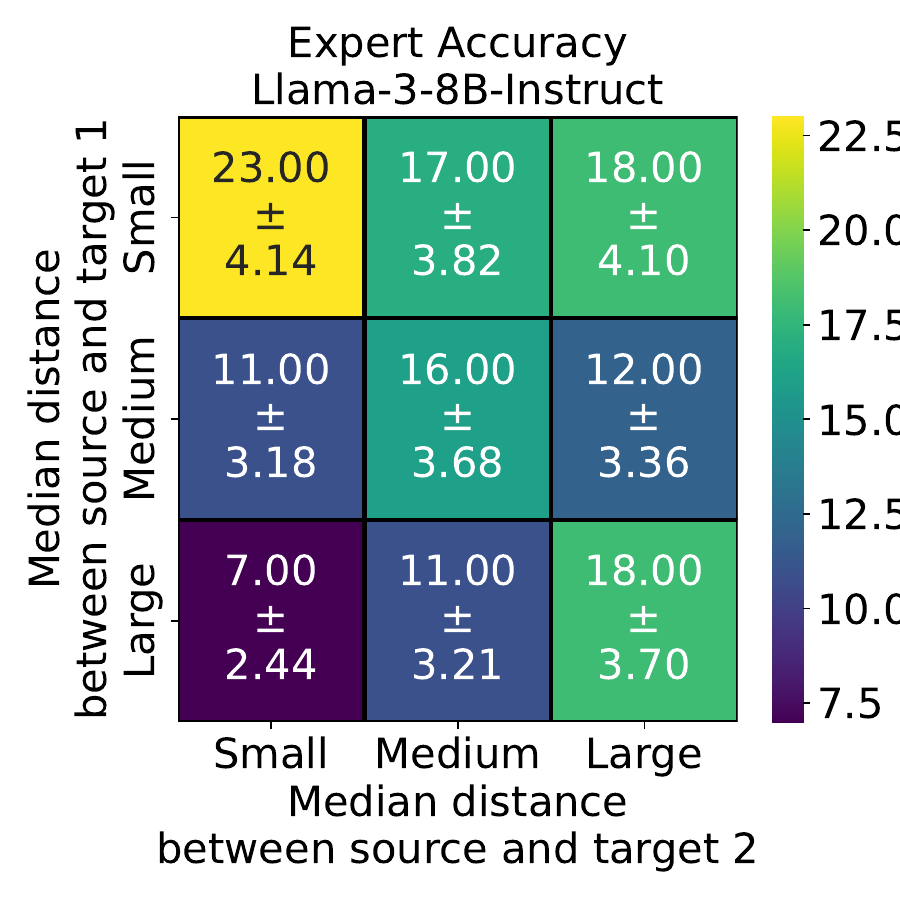}
    \end{subfigure}

        \begin{subfigure}{0.33\linewidth}
        \centering
        \includegraphics[width=\linewidth]{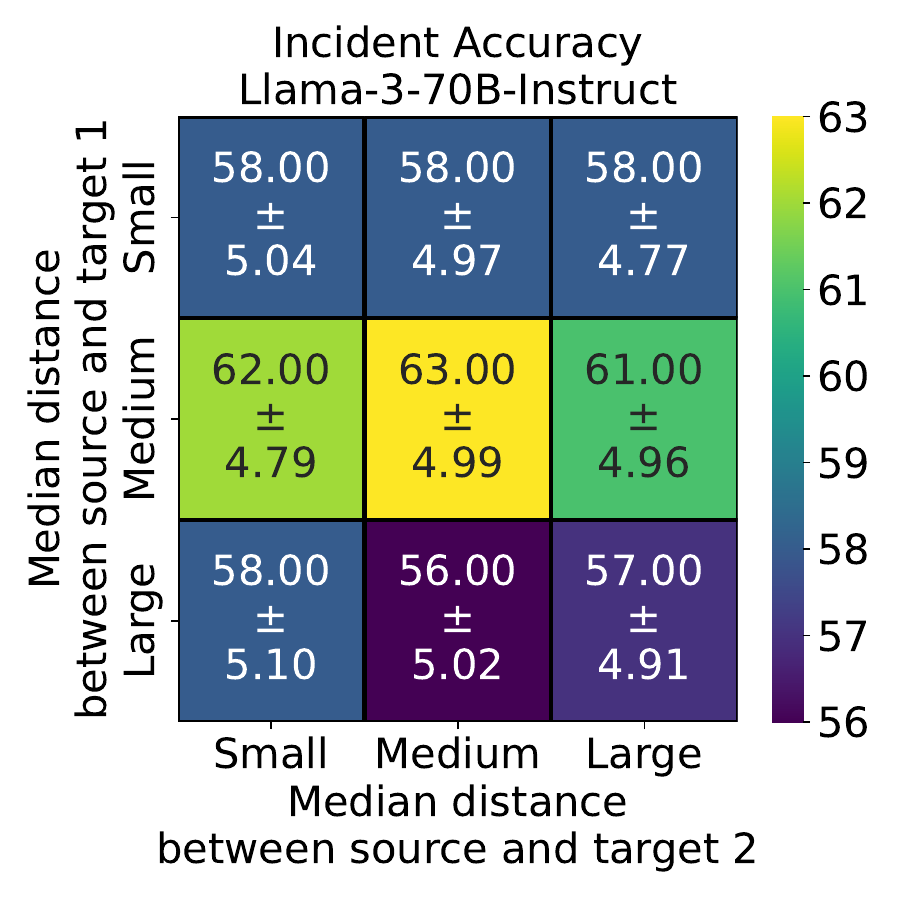}
    \end{subfigure}
    \hspace{-0.02\linewidth}
    \begin{subfigure}{0.33\linewidth}
        \centering
        \includegraphics[width=\linewidth]{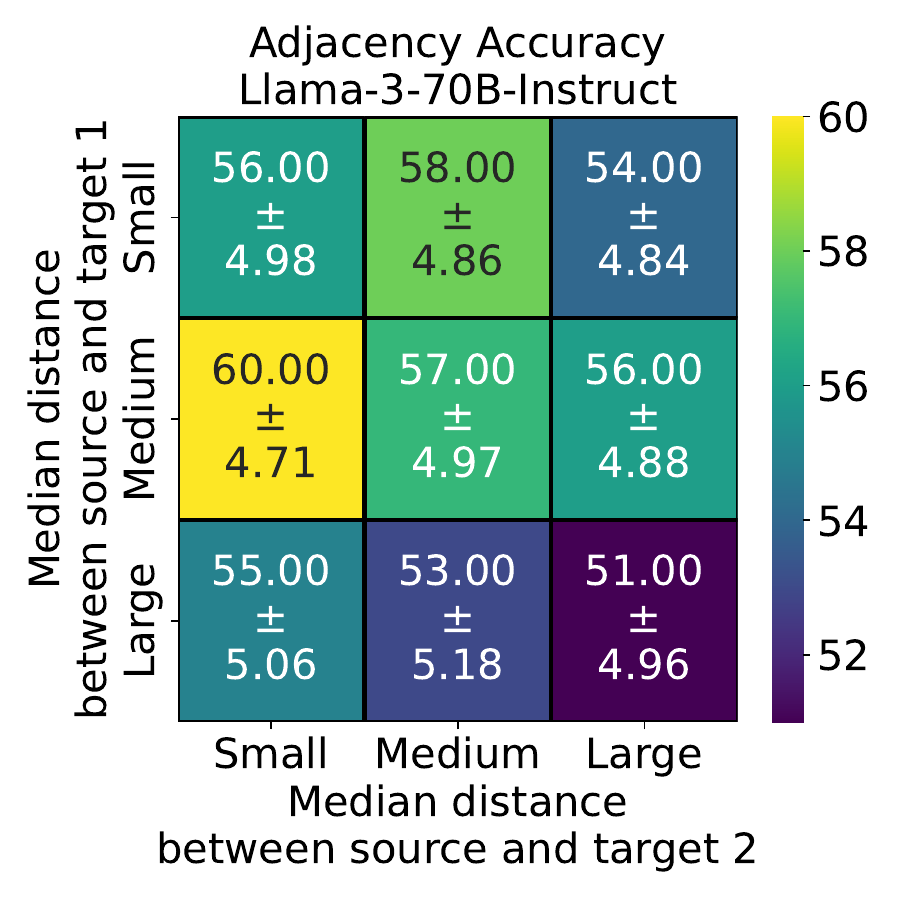}
    \end{subfigure}
    \hspace{-0.02\linewidth}
    \begin{subfigure}{0.33\linewidth}
        \centering
        \includegraphics[width=\linewidth]{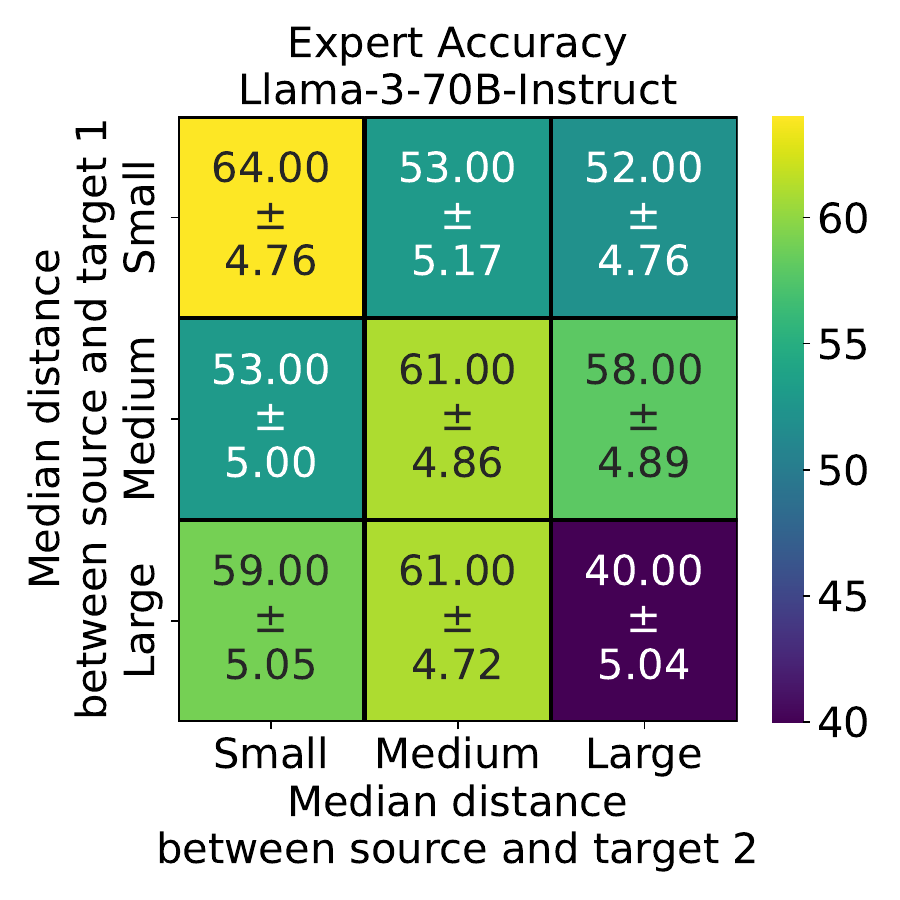}
    \end{subfigure}
    \caption{All results in the similarity task with density = 0.1.}
    \label{all_result_similarit_fig}
\end{figure}

Figure~\ref{all_result_similarit_fig} presents the results of the similarity task at a density of 0.1 across three models (GPT-4, Llama-8B, Llama-70B) and three different graph encodings. For all models using the graph encoding functions, we observe the typical lost-in-distance pattern, where performance at the $(\mathrm{Large}, \mathrm{Large})$ index—indicating a large distance between relevant information within two subgraphs in the prompt—is worse than at the $(\mathrm{Small}, \mathrm{Small})$ index, which indicates a smaller distance between relevant information for the similarity task in the prompt.

\section{Effect of Graph Density}\label{s:appendix:impact_of_seq_le}
The lost-in-distance effect remains consistent across different graph densities, i.e., different values of $P(e_{ij} \in \gE)$ in Erdős–Rényi (ER) randomly generated graphs. Graph density affects the input sequence length in a linear manner; higher densities result in proportionally longer input sequences, as demonstrated in Table~\ref{t:graph_density_tokens}. 

\begin{figure}[ht]
    \centering
    \begin{subfigure}{0.33\linewidth}
        \centering
        \includegraphics[width=\linewidth]{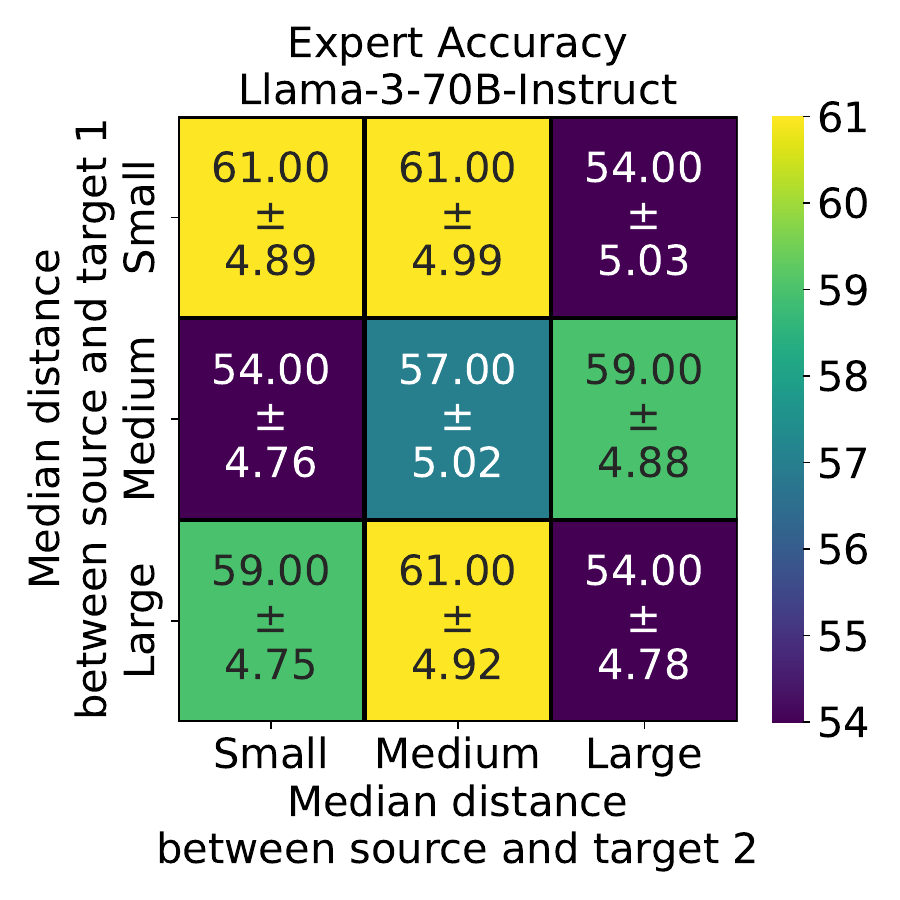}
    \end{subfigure}
    \hspace{-0.02\linewidth}
    \begin{subfigure}{0.33\linewidth}
        \centering
        \includegraphics[width=\linewidth]{./Figures/similarity_median/Llama-3-70B-Instruct-expert.pdf}
    \end{subfigure}
    \hspace{-0.02\linewidth}
    \begin{subfigure}{0.33\linewidth}
        \centering
        \includegraphics[width=\linewidth]{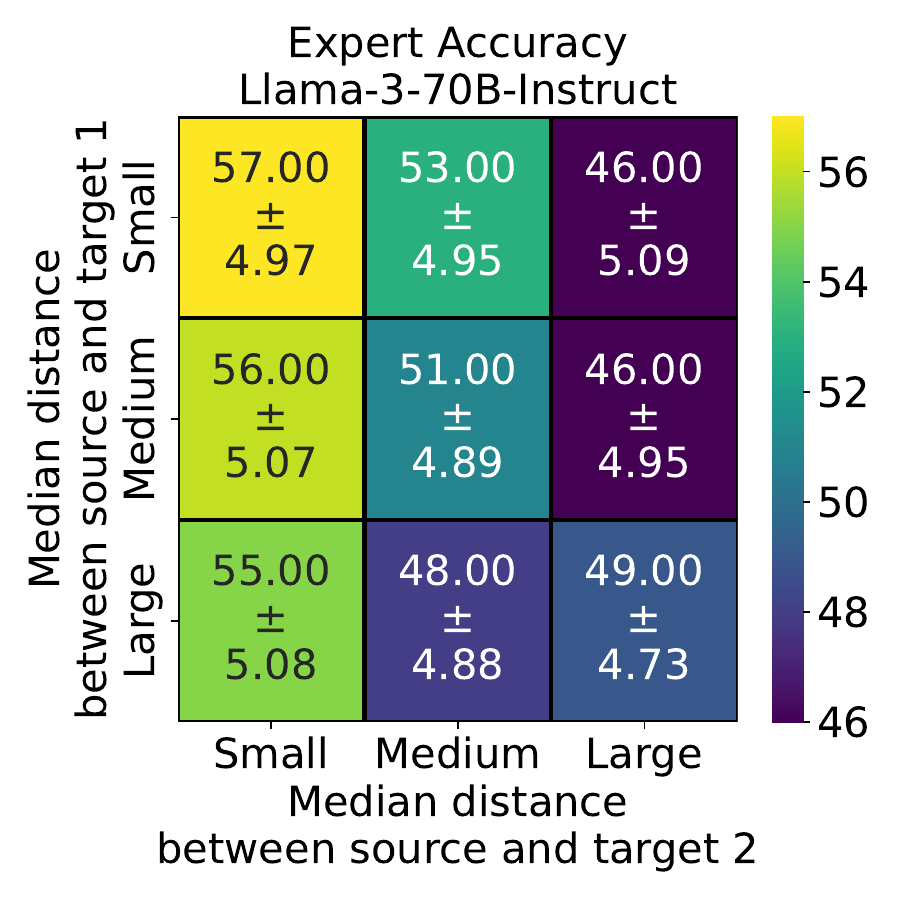}
    \end{subfigure}
    \caption{Results from similarity tasks with three different density values, $P(e_{ij} \in \gE)$, (left) 0.05 , (middle) 0.1, and (right) 0.15.}
    \label{similarity_sequence_length_fig}
\end{figure}

Figure~\ref{similarity_sequence_length_fig} illustrates that increasing the context length by raising graph density follows the same pattern of the lost-in-distance effect in similarity tasks. Specifically, accuracy declines progressively from top to bottom and left to right as the distances between common edges within each subgraph increase. Additionally, the figure demonstrates that for smaller context lengths, corresponding to graphs with low density, the results are noisier and the effect of lost-in-distance diminishes. 

\begin{table}[ht]
    \centering
    \caption{Average Token Length from different Graph Density, $P(e_{ij} \in \gE)$, for ER graphs using expert encoding}
    \begin{tabular}{cc}
        \hline
        \textbf{Graph Density} & \textbf{Average Token Length} \\
        \hline
        0.05 & 1145 \\
        0.10 & 1869 \\
        0.15 & 2633 \\
        \hline
    \end{tabular}
    \label{t:graph_density_tokens}
\end{table}

\end{document}